\documentclass{article}

\PassOptionsToPackage{numbers, compress}{natbib}

\usepackage[final]{neurips_2025}
\usepackage{xspace}

\usepackage{color}
\usepackage{colortbl}
\usepackage{xcolor}
\definecolor{citecolor}{HTML}{0071BC}
\definecolor{linkcolor}{HTML}{ED1C24}

\usepackage[utf8]{inputenc} 
\usepackage[T1]{fontenc}    
\usepackage[pagebackref, breaklinks=true, colorlinks,citecolor=citecolor, linkcolor=linkcolor, bookmarks=false]{hyperref}
\usepackage{url}            
\usepackage{booktabs}       
\usepackage{amsfonts}       
\usepackage{nicefrac}       
\usepackage{microtype}      
\usepackage{xcolor}         

\usepackage{fancyvrb}   
\usepackage{listings}   
\usepackage{multirow}
\usepackage{algorithm}
\usepackage[noEnd=false,indLines=true]{algpseudocodex}

\usepackage{bm}
\usepackage{bbm}
\usepackage{amsmath}
\usepackage{wrapfig}
\usepackage{graphicx}
\usepackage{xcolor} 
\usepackage{colortbl}  
\usepackage{array}     
\usepackage[utf8]{inputenc}

\definecolor{mycolor_blue}{HTML}{FFFFFF}
\definecolor{mycolor_green}{HTML}{EEEEEE}
\definecolor{mycolor_gray}{HTML}{FFFFFF}
\definecolor{pearDark}{HTML}{2980B9}

\usepackage{tcolorbox}
\usepackage{fbox}
\definecolor{textcolor1}{rgb}{0.25,0.5,0.5}
\definecolor{textcolor2}{rgb}{0.7,0.25,0.25}
\definecolor{linkc}{rgb}{0, 0.44, 0.74}
\definecolor{eqc}{rgb}{1, 0, 0}
\definecolor{myy}{RGB}{126,95,0}
\definecolor{mygray}{gray}{.9}
\definecolor{bblue}{RGB}{30,80,120}
\definecolor{mygray1}{gray}{.7}
\definecolor{ggray}{RGB}{127,127,127}
\definecolor{mygreen}{RGB}{93,174,86}
\definecolor{scolor}{RGB}{111,168,220}
\definecolor{hcolor}{RGB}{111,176,81}
\definecolor{ocolor}{RGB}{224,103,102}
\definecolor{wcolor}{RGB}{246,178,107}
\definecolor{citecolor}{HTML}{229954}
\definecolor{linkcolor}{RGB}{237,4,140}

\def\methodNAME{InfinityStar\xspace}

\title{\methodNAME: Unified Spacetime AutoRegressive Modeling for Visual Generation}

\vspace{-5pt}
\author{
  \vspace{-25pt}\\
  \textbf{Jinlai Liu\thanks{Equal contribution. $^{\dag}$Corresponding author: \href{mailto:yuanzehuan@bytedance.com}{\color{black}{yuanzehuan@bytedance.com}}}$~~$,\quad Jian Han$^{*}$,\quad Bin Yan$^{*}$}\vspace{3pt} \textbf{\quad Hui Wu, \quad Fengda Zhu,\quad Xing Wang}\vspace{3pt} \\
  \textbf{\quad Yi Jiang,\quad Bingyue Peng,\quad Zehuan Yuan$^{\dag}$}\vspace{3pt} \\
  ByteDance\vspace{3pt} \\
  \texttt{\small \{liujinlai.licio,hanjian.thu123,bin.yan,wuhui.321,fengdazhu\}@bytedance.com,}\\ 
  \texttt{\small \{xing.wang,jiangyi.enjoy,bingyue.peng,yuanzehuan\}@bytedance.com,}\vspace{8pt}  \\
  Codes and models:~\, \url{https://github.com/FoundationVision/InfinityStar}
  \vspace{-4pt} \\
}

\begin{document}

\maketitle

\begin{abstract}

We introduce \methodNAME, a unified spacetime autoregressive framework for high-resolution image and dynamic video synthesis. Building on the recent success of autoregressive modeling in both vision and language, our purely discrete approach jointly captures spatial and temporal dependencies within a single architecture. This unified design naturally supports a variety of generation tasks such as text-to-image, text-to-video, image-to-video, and long interactive video synthesis via straightforward temporal autoregression. Extensive experiments demonstrate that \methodNAME scores 83.74 on VBench, outperforming all autoregressive models by large margins, even surpassing some diffusion competitors like HunyuanVideo. Without extra optimizations, our model generates a 5s, 720p video approximately 10$\times$ faster than leading diffusion-based methods. To our knowledge, \methodNAME is the first discrete autoregressive video generator capable of producing industrial-level 720p videos. We release all code and models to foster further research in efficient, high-quality video generation.

\end{abstract}
    
\section{Introduction}
\label{sec:intro}

Visual synthesis has witnessed remarkable progress in recent years, largely propelled by the scaling of Transformer architectures. In particular, video generation has attracted growing interest from both academia and industry, owing to its wide-ranging applications in content creation, world simulation, etc. At present, diffusion models\cite{sora,kling,Hunyuanvideo,Wan,veo3,waver} lead the field by iteratively denoising latent representations to produce high-fidelity clips. Concurrently, autoregressive models\cite{videopoet,wang2024emu3,nova} have been explored for their potential to unify image and video generation and to generalize over longer time horizons.

Despite their successes, each paradigm exhibits critical shortcomings. Video diffusion models excel at synthesizing fixed‐length frame sequences by exploiting bidirectional attention, yet they incur substantial computational cost due to tens or even hundreds of sequential denoising steps, and they struggle to extend seamlessly to video extrapolation. Autoregressive methods based on next-token prediction, while inherently capable of streaming generation, often fall short in visual fidelity and suffer from prohibitive latency due to tens of thousands of inference steps.

These observations motivate the need for a generation framework that simultaneously possess high visual quality, efficiency and temporal generalization. Recently, Visual AutoRegressive modeling (VAR)\cite{VAR} redefined image generation as a coarse-to-fine next-scale prediction. Its follow-up work, Infinity~\cite{Infinity} further introduces bitwise modeling and scales up the vocabulary size, achieving comparable performance to diffusion models while offering significant advantages in inference speed. Inspired by the success of VAR~\cite{VAR} and Infinity~\cite{Infinity}, we present \methodNAME, a Spacetime Pyramid Modeling for unified text‐to‐image, text‐to‐video, zero-shot image‐to‐video, and zero‐shot video extrapolation. This framework models a video as an image pyramid and multiple clip pyramids, not only naturally inheriting the text-to-image capabilities but also decoupling static appearance from dynamic motions in videos. Furthermore, we introduce several key improvements. First, we improve discrete reconstruction quality by leveraging knowledge inheritance from a continuous video tokenizer. Second, we introduce Stochastic Quantizer Depth during training of the tokenizer to alleviate the imbalanced information distribution across scales. Third, we propose Semantic Scales Repetition, which refines the predictions of early semantic scales in a video, significantly enhancing fine-grained details and complex motions of the generated videos.

We train \methodNAME on large‐scale video corpora to support generating videos of up to 720p resolution and variable durations. On the VBench benchmark\cite{vbench}, \methodNAME establishes a new state-of-the-art among autoregressive video generation models, even surpassing industry-leading HunyuanVideo\cite{Hunyuanvideo} (83.74 v.s 83.24). Besides, \methodNAME shows a great advantage in terms of speed. Using visual tokenizers of the same compression rate, \methodNAME achieves a $10\times$ reduction in inference latency relative to leading diffusion models.

In summary, the main contributions of our work are as follows:
\begin{enumerate} 
\vspace{-1pt}
\item We propose \methodNAME, a novel spacetime pyramid modeling framework that unifies diverse visual generation tasks, demonstrating superior flexibility and versatility.
\vspace{-1pt}
\item \methodNAME is the first discrete autoregressive model capable of generating high-quality videos, outperforming existing autoregressive text-to-video models and matching the performance of leading diffusion models.
\vspace{-1pt}
\item Compared to the inefficiency of existing autoregressive models and diffusion models, \methodNAME significantly accelerates high-quality video generation.
\end{enumerate}

\section{Related Work}
\label{sec:related_works}

\subsection{Video Diffusion Models}
Diffusion models excel at generating high-fidelity data by gradually denoising random noise and has been widely applied in video generation. Early attempts~\cite{svd,videocrafter,pixeldance} are built on U-Net architectures, demonstrating the feasibility of this approach but falling short in producing sharp, temporally coherent frames due to limited model capacity. The advent of Diffusion Transformers (DiT~\cite{dit}) marked a turning point. SORA~\cite{sora} harnessed DiT’s scaling ability to process spatio-temporal patches at scale, dramatically improving both video consistency and generation quality. The success of SORA has catalyzed a wave of innovation~\cite{cogvideox,Hunyuanvideo,Wan,waver} across the industry, propelling video generation to unprecedented levels of coherence and fidelity. Although video diffusion models deliver outstanding quality, their slow generation speed hinders the production of high-resolution, long-duration videos.

\subsection{Video AutoRegressive Models}
Another class of methods~\cite{wang2024emu3,nova,videopoet} employs autoregressive models for video generation. Inspired by the success of LLMs, these works predict video tokens in specific orders using an autoregressive Transformer. For example, Emu3~\cite{wang2024emu3} performs next-token prediction along both spatial and temporal axes, while Nova~\cite{nova} first predicts spatial tokens set-by-set and subsequently proceeds frame-by-frame in the temporal dimension. Although achieving preliminary progress, they require hundreds to thousands of inference steps, resulting in prohibitively low generation efficiency. In contrast, recent advances in next-scale prediction~\cite{VAR,Infinity} have demonstrated state-of-the-art performance in image synthesis, offering both improved quality and markedly faster inference. In this work, we extend the next-scale prediction paradigm to the unified image and video generation.

\subsection{Discrete Video Tokenizers}
For a long time, discrete~\cite{videogpt,magvit2} and continuous~\cite{Hunyuanvideo,cogvideox,Wan} video tokenizers have been developed independently. Although some works~\cite{cosmos,omnitokenizer} provide both discrete and continuous tokenizers, the network configurations are usually not aligned. For example, Cosmos~\cite{cosmos} chooses 6 and 16 as latent dimensions in its discrete and continuous variants respectively. This misalignment hinders the knowledge reuse between two types of tokenizers. As a result, most mainstream discrete video tokenizers are either trained from scratch~\cite{cosmos} or starting from a pretrained discrete image tokenizer~\cite{magvit2,omnitokenizer}. However, these training strategies have the following drawbacks. First, training from scratch is inefficient and converges slowly. Second, weights pretrained on static images are not optimal for video reconstruction. 
To alleviate these deficiencies, we propose a new training strategy, which inherits the architecture and knowledge of a trained continuous video tokenizer. Experiments show that this strategy significantly boosts the convergence of discrete video tokenizers.

\section{\methodNAME Architecture}

\subsection{Preliminaries}
\label{sec:method-preliminaries}

\noindent\textbf{Infinity for Image Generation.}
Infinity~\cite{Infinity} decomposes an image into a sequence of hierarchical token blocks using a visual tokenizer and models the relationship between tokens by a visual autoregressive Transformer (VAR Transformer). To cover images of various sizes, Infinity pre-defines a list of token block sizes $\{(h_1,w_1),...(h_K,w_K)\}$, called scale schedule. The size $(h_i,w_i)$ in scale schedule grows as $i$ increases, forming a pyramid-like structure, which we refer as \textbf{image pyramid} in later discussion. Next we introduce the training and inference procedure of Infinity.

In the first training stage, a visual tokenizer learns to reconstruct the raw image and compress it into a sequence of discrete tokens, which can be modeled by the VAR Transformer in the next stage. Specifically, the tokenizer first encodes the raw images into compact latents, then transforms latents into $K$ discrete residual token blocks $(r_1, r_2, ..., r_K)$ using a bitwise multi-scale residual quantizer~\cite{Infinity}. Each token block $r_i$ consists of $h_i \times w_i$ discrete tokens of $d$-dim with vocabulary size of $2^d$. Then in the second stage, a VAR Transformer is trained to predict next residual token block $r_k$ conditioned on text embedding $\psi(t)$ and former tokens blocks $r_{<k}$. Formally, in each step, VAR Transformer predicts a conditional probability $p(r_k|r_{<k},\psi(t))$. During the inference, Infinity generates an image by running the VAR Transformer $K$ times autoregressively, merging the predicted tokens and running the tokenizer decoder once.

\begin{figure}[t]
  \centering
   \includegraphics[width=1.0\linewidth]{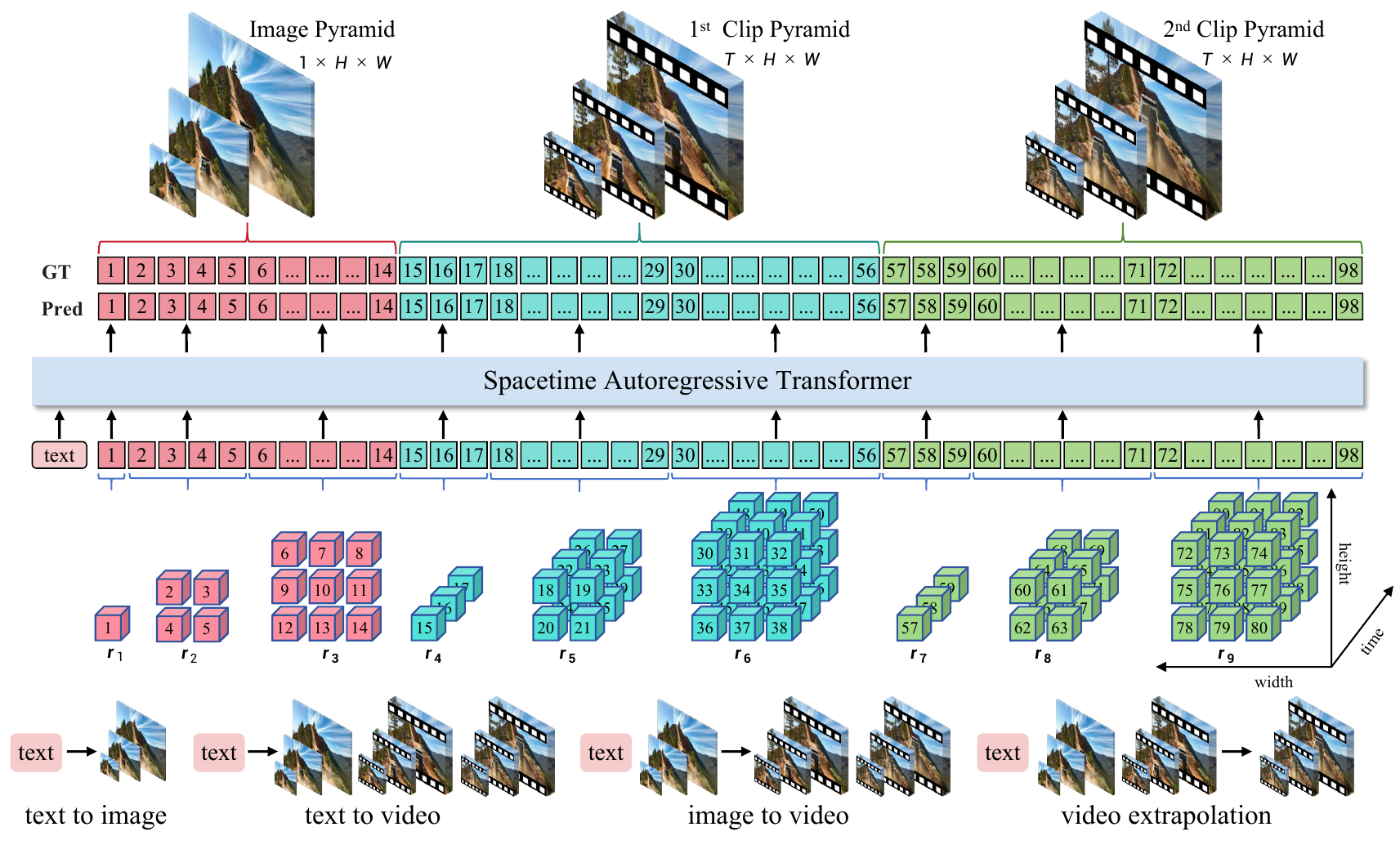}
   \vspace{-0.7cm}
   \caption{\textbf{Spacetime pyramid modeling of \methodNAME.} Built with an unified autoregressive pipeline,  \methodNAME is capable of performing text-to-image, text-to-video, image-to-video, video extrapolation tasks all in one model.}
   \label{fig:framework}
\end{figure}

\subsection{Spacetime Pyramid Modeling for Unified Generation}
\label{sec:method-modeling}

Extending the spatial-only next-scale prediction paradigm of Infinity~\cite{Infinity} to video generation presents a primary challenge: \textit{how to incorporate the temporal dimension}. The straightforward strategies are either letting time grows uniformly, \emph{i.e.}, from $(1,1,1)$ to $(T,H,W)$, or keeping time constant, \emph{i.e.}, from $(T,1,1)$ to $(T,H,W)$. We empirically found that letting time grow uniformly produces flickering videos. As for the constant time pyramid, we refer to it as the \textbf{pseudo-spacetime pyramid}. Despite its conceptual simplicity, it suffers from two fundamental limitations. First, the treatment of videos differs markedly from that of images, preventing a text-to-video (T2V) model from effectively leveraging the knowledge learned by a text-to-image (T2I) model and complicating its extension to tasks such as image-to-video (I2V). Second, because appearance and motion in videos are coupled in this design, the model faces significant difficulty in accurately fitting both aspects.

To overcome these challenges, we propose a novel \textbf{spacetime pyramid modeling} framework as shown in Fig.\ref{fig:framework}. Each video is decomposed into sequential clips $\{c_1,c_2,\cdots,c_N\}$. {We regard the first frame as $c_1$ (\emph{i.e.}, $T=1$) to specifically encode static appearance cues in videos and other clips share an equal duration $T>1$}. Each clip is modeled as a 3D volume pyramid similar to Infinity~\cite{Infinity}. In particular, for each clip, there are $K$ scales with each represented as a residual token block $r_k$ of $(T, h_k, w_k)$ dimension. \emph{It is worth noting that all scales in the pyramid are extended only in spatial dimension instead of time}. Mathematically, the tokens in the first clip are generated auto-regressively across scales as: 
\vspace{-0.1cm}
\begin{equation}
p(r_1^1, \dots, r_K^1) = \prod_{k=1}^{K} p(r_k^1 \mid r_1^1, \dots, r_{k-1}^1, \psi(t)),
\end{equation}
For inter-clip predictions, clips are generated sequentially conditioned on prior clip predictions and the text input in an autoregressive manner. In this way, we could generate infinitely long videos theoretically. Formally, the autoregressive likelihood of the whole video can be expressed as:
\begin{equation}
p(r_1^1, \dots, r_K^N) = \prod_{c=1}^{N}\prod_{k=1}^{K} p(r_k^c \mid r_1^1, \dots, r_{k-1}^c, \psi(t)),
\end{equation}

\begin{figure}[t]
  \centering
   \includegraphics[width=0.95\linewidth]{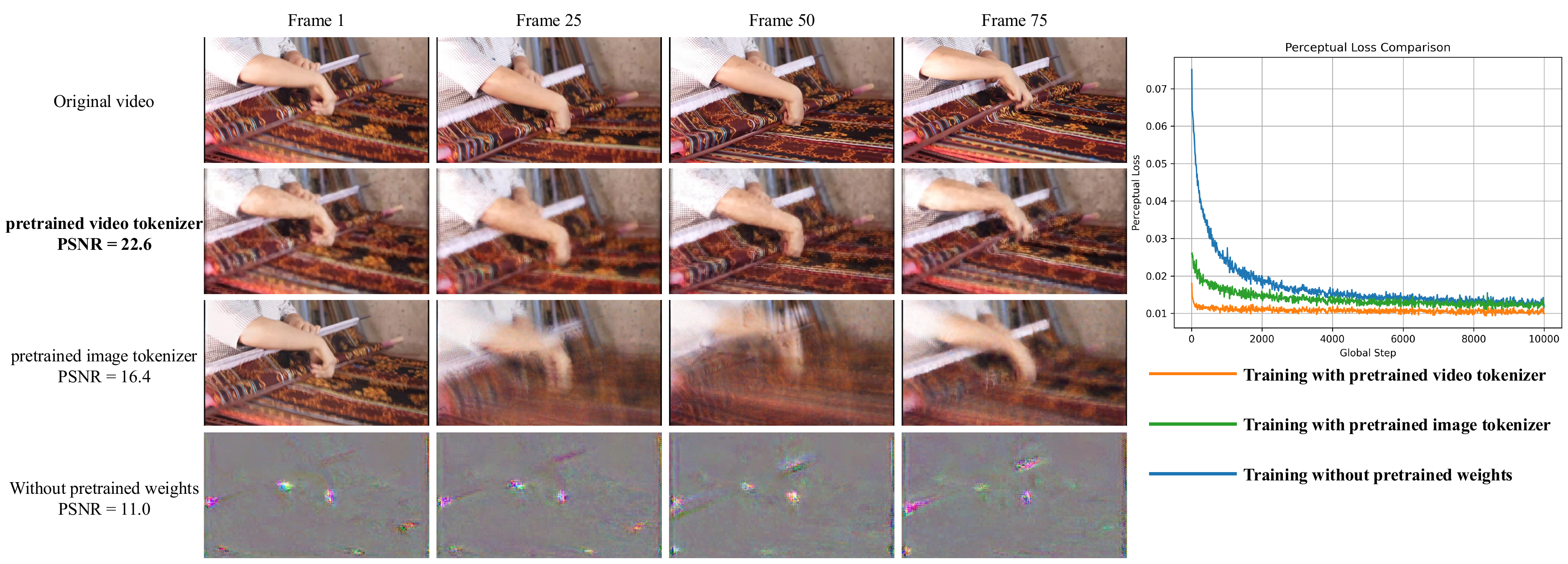}
   \vspace{-0.2cm}
   \caption{Influence of pretrained weights on reconstruction and convergence. The left sub-figure shows the reconstructed frames using different pretrained weights \texttt{without finetuning}. Loading weights of continuous video tokenizer achieves the best results. The right sub-figure shows that training with pretrained video tokenizer converges significantly faster than the other two strategies. }
   \label{fig:recon-compare}
   \vspace{-0.2cm}
\end{figure}

\begin{figure}[t]
  \centering
   \includegraphics[width=0.9\linewidth]{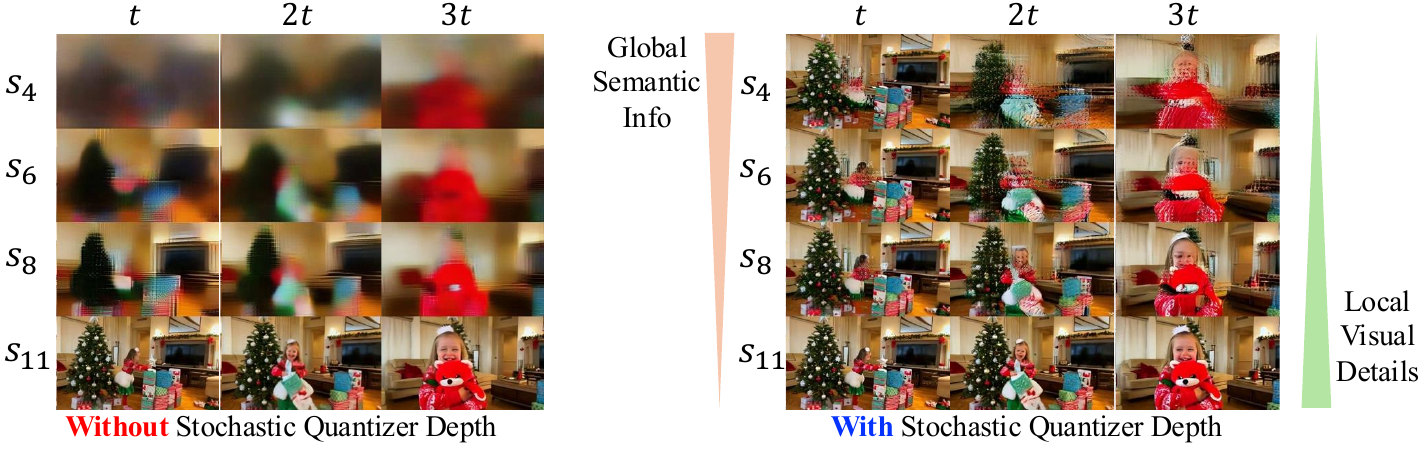}
   \vspace{-0.3cm}
   \caption{The influence of stochastic quantizer depth. Sub-figure $(s_i, nt)$ represents the reconstructed frame $nt$ using all tokens from the image pyramid plus tokens of first $i$ scales in the clip pyramid. SQD significantly improves the reconstruction quality of early scales. Besides, the earlier scales correspond to global semantics, while the later ones are responsible for local visual details.}
   \label{fig:sqd_recon_vis}
   \vspace{-0.4cm}
\end{figure}

\subsection{Visual Tokenizer}
\label{sec:method-tokenizer}

Training video tokenizers faces greater challenges than training image tokenizers. First, training tokenizers on videos of tens of frames is much computationally heavier than training on static images. Therefore, training a video tokenizer from scratch is extremely time-consuming and suffers from slow convergence. Second, the scale schedule in videos leads to more imbalanced information distribution, where most information is concentrated in the last few scales. This brings great difficulties to the optimization of VAR Transformer. To solve these challenges, we introduce two techniques, knowledge inheritance from continuous video tokenizer and stochastic quantizer depth. 

\noindent \textbf{Knowledge Inheritance from Continuous Video Tokenizer}. Instead of designing and training a discrete video tokenizer from scratch, we inherit the architecture and weights of a trained continuous video tokenizer, \emph{i.e.} video VAE. Specifically, we first insert a parameter-free quantizer between the pre-trained VAE encoder and the decoder. The quantizer is based on binary spherical quantization~\cite{BSQ}, being similar to that of Infinity~\cite{Infinity} but with new spacetime pyramid scale schedule. This does not introduce any new parameter like codebook in VQ~\cite{vqvae} and well retains knowledge of the original VAE. As shown in Fig.\ref{fig:recon-compare}, the discrete video tokenizer reconstructs videos decently, even without any fine-tuning. To further improve the reconstruction quality, we fine-tune the new tokenizer jointly on images and videos like previous works~\cite{omnitokenizer,cosmos}. During the fine-tuning, the KL loss of the original VAE is replaced with the commitment loss plus the entropy penalty~\cite{BSQ}. As shown in Fig.\ref{fig:recon-compare}, with the help of knowledge of continuous video VAE, the convergence accelerates dramatically.

\noindent \textbf{Stochastic Quantizer Depth}. When tokenizing videos using the spacetime pyramid schedule, the information distribution on different scales gets extremely imbalanced. Specifically, there are only a few tokens in the early scales, while there are tens of thousands of tokens in the last scales. Thus the tokenizer tends to reconstruct videos solely relying on tokens from the last few scales and not to learn useful representation in early scales as shown in Fig.\ref{fig:sqd_recon_vis} (left). However, this imbalanced distribution is difficult to model using VAR Transformer because the dependence between the latter token blocks and the former ones is weak. To alleviate this problem, we propose a regularization called stochastic quantizer depth. During training, each one of the last $N$ scales has a probability $p$ of being discarded. In this way, there are $2^{N}$ possible scale schedules during training. This requires the tokenizer to reduce the reliance on last scales and store more information in tokens of early scales. As in Fig.\ref{fig:sqd_recon_vis} (right), with the help of this regularization, the reconstruction results of early scales become much clearer. This balanced information distribution makes the training of VAR Transformer easier.

\subsection{Spacetime Autoregressive Transformer}
\label{sec:method-transformer}
To accommodate the newly introduced temporal dimension, enhance the quality of generated videos, and alleviate the substantial computational overhead associated with a large number of tokens, we propose the following modifications to the VAR Transformer: Semantic Scale Repetition, Spacetime Sparse Attention, and Spacetime RoPE. We put Spacetime RoPE in the appendix~\ref{appendix:A}.

\noindent\textbf{Semantic Scale Repetition.} With carefully crafted positional encodings, \methodNAME can already generate videos of acceptable quality. However, we observe that the structural coherence and motion dynamics in these outputs remain suboptimal. As shown in Figure~\ref{fig:sqd_recon_vis}, the overall layout and the placement of foreground objects are determined by the early scales of the clip pyramid—what we term the ``semantic scales.'' This observation motivates us to enhance generation fidelity at these semantic scales. To this end, we introduce a simple yet effective technique called semantic scale repetition. Concretely, if a clip pyramid comprises 
$K$ scale tuples, we repeat the first $K_s$ tuples $N$ times, thereby reinforcing the semantic representations. In this way, every early residual $r_k$ undergoes multiple rounds of refinement, improving the generation quality of semantics and the performance in complex scenarios with large motion. Given that the tokens at these early scales account for only a small fraction of the total token count, the additional computational overhead incurred by repeating them is negligible.

\begin{figure}[t]
    \vspace{-0.2cm}
  \centering
   \includegraphics[width=1\linewidth]{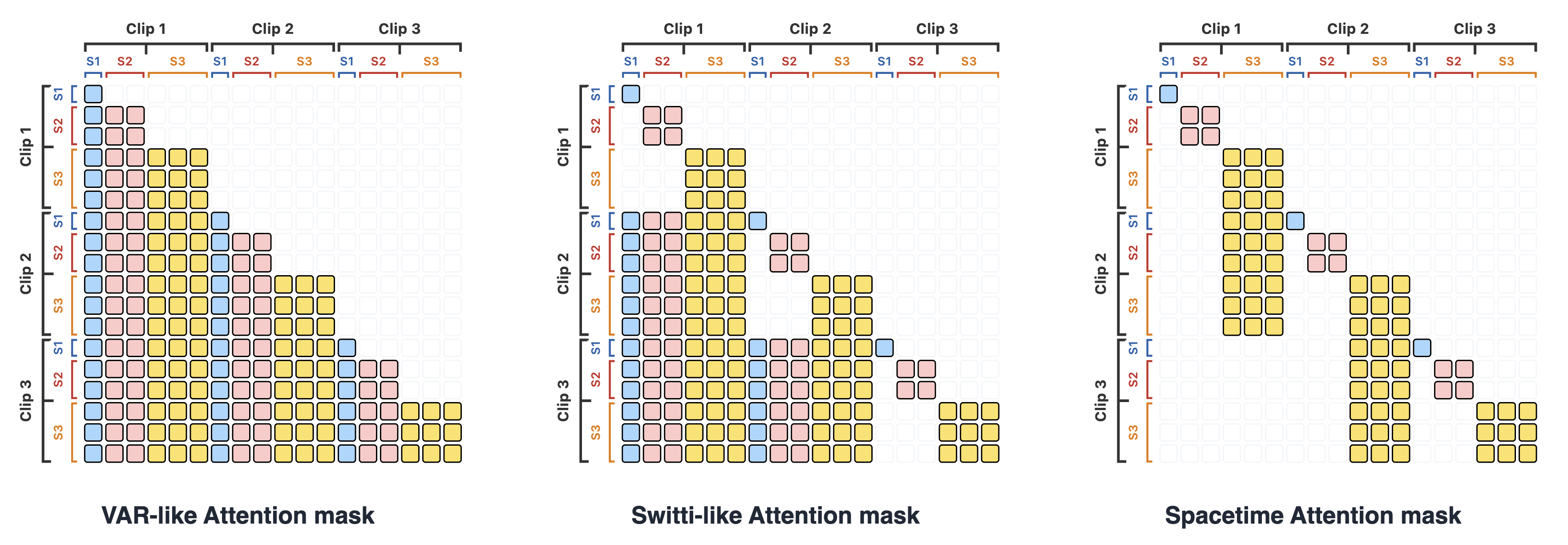}
   \vspace{-0.5cm}
   \caption{Illustration of three causal attention variants. We plot three pyramids on the scale size = (1,2,3) for visualization simplicity.  From left to right,  VAR block-wise causal mask with full history,  Switti block-wise non-causal mask with full history,  and spacetime sparse attention.}
   \label{fig:sparse_attention}
   \vspace{-0.2cm}
\end{figure}

\noindent\textbf{Spacetime Sparse Attention.} Autoregressive video generation faces significant challenges due to the high computational costs of long context. As on the left of Fig.\ref{fig:sparse_attention}, Infinity~\cite{Infinity} employs a block-wise causal mask for single pyramid modeling. Switti~\cite{Switti} verifies that conditioning solely on inputs from preceding scales is sufficient for next-scale predictions, resulting in a sparser attention mask as on the middle of Fig.\ref{fig:sparse_attention}. For long video generation, it's necessary to attend history tokens to achieve temporal consistency. However, attending full history leads to an explosively long sequence. Considering each clip corresponds to 5s, which is sufficient to maintain temporal consistency, here we only attend to the last scale of the preceding clip. Finally, we obtain a highly sparse attention as show in Fig.\ref {fig:sparse_attention} (right). Our spacetime sparse attention drastically reduces computational overhead in attention during both training and inference, while delivering better performance.

\renewcommand{\algorithmicrequire}{\textbf{Input:}}
\renewcommand{\algorithmicensure}{\textbf{Output:}}

\label{sec:experiment}

\begin{figure}[t]
  \centering
   \includegraphics[width=1.0\linewidth]{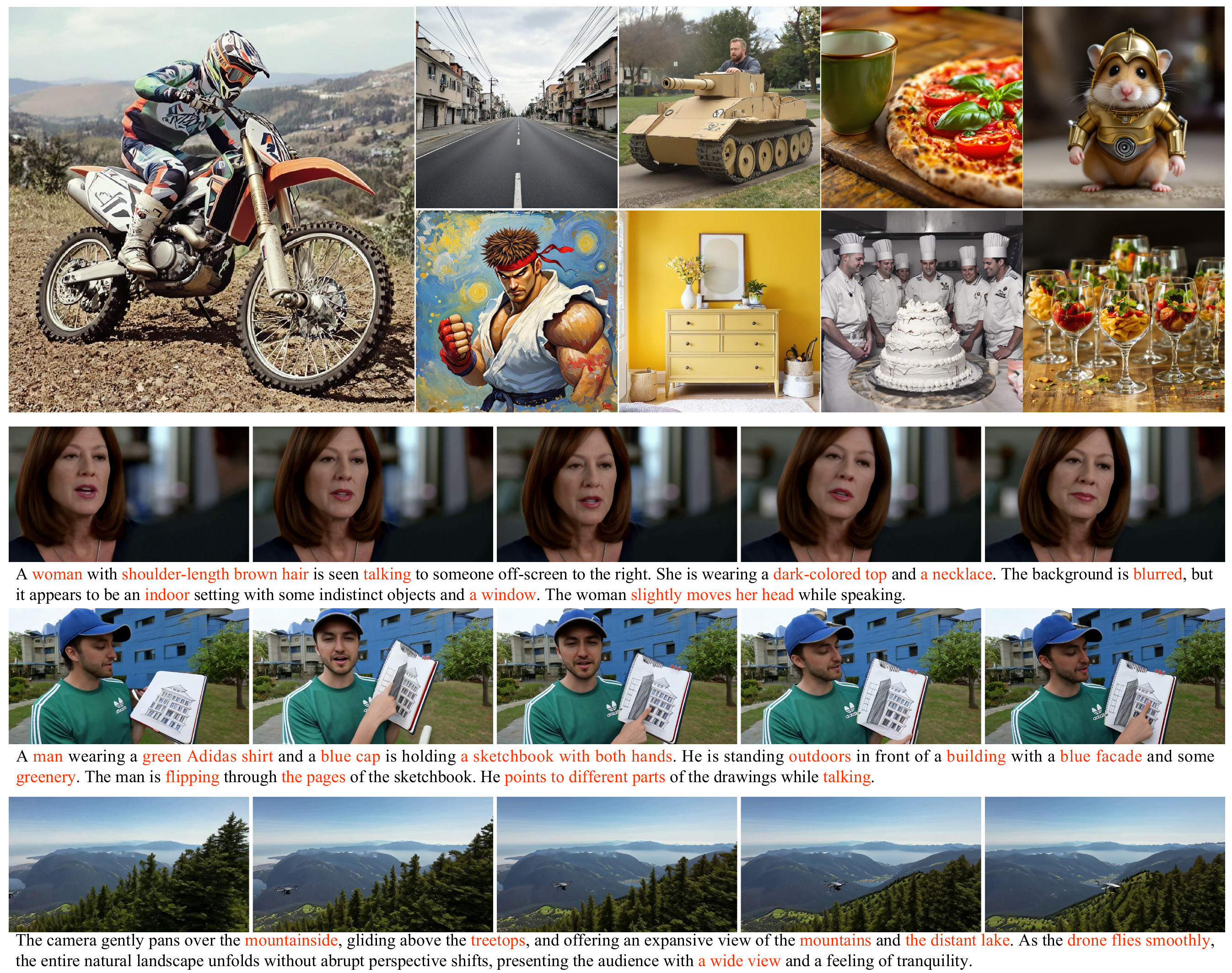}
   \vspace{-0.5cm}
   \caption{Text to image and text to video examples.}
   \label{fig:show_cases}
   \vspace{-0.5cm}
\end{figure}

\section{Experiment}

\subsection{Implementation}

\textbf{Datasets.} The training data of \methodNAME includes text-to-image data and text-to-video data. We curated 130M pretraining and 70M high-quality text-to-image data. To balance the data distribution and improve overall aesthetics, we also involve 5M high-quality synthetic data. In terms of text-to-video data, we curated around 16M video data. All videos are longer than 5 seconds. Among them 13M videos are under 336$\times$192 resolution used for pre-training. They are mainly from Panda-70M\cite{panda70m}, Mira\cite{miradata}, and other internal video-text pairs. Apart from those 192p videos, we also curated 3M 480p and 50K 720p high-quality videos for fine-tuning.

\textbf{Model and Training.} After inserting the patchify and unpatchify layers between Wan 2.1 VAE's encoder and decoder, we obtain a video tokenizer with a compression rate of $4\times16\times16$ and a latent dimension of 64. Multi-scale BSQ quantization is adopted to obtain discrete tokens. In contrast to using a vocabulary size of $2^{64}$ for all scales, we use a vocabulary size of $2^{16}$ for the former small scales and $2^{64}$ for the latter large scales. We empirically find that it boosts convergence and has a negligible impact on the reconstruction quality. Starting with the pretrained weights of Wan 2.1 VAE, the discrete tokenizer is fine-tuned jointly on images of $256\times256$, $512\times512$, $768\times768$ and videos of $256\times256\times81$ for 30K iterations. The learning rate is $1e^{-4}$.

The autoregressive Transformer of \methodNAME is trained progressively in four stages, including a T2I pre-training and three T2V fine-tuning on 192p, 480p, 720p respectively. Each time we increase the training resolution, we preserve scale schedule of lower resolutions and append several larger scales, which enables better inheritance. The global batch size for 192p is 2048 and that of 480p and 720p is 1024. The learning rate for 192p is $2e^{-4}$. Then we decay it to $1e^{-4}$ for 480p and 720p. We train the model on videos of 192p, 480p, 720p for 50K, 8K, 3K iterations, respectively. Specifically, each clip pyramid is composed of 80 frames at 16 fps, and the first $K_s=12$ semantic scales are repeated by $N=3$ times. Details about infrastructure optimizations are presented in the appendix~\ref{appendix:B}.

\subsection{Text-to-Image Generation}
The upper part of Fig.\ref{fig:show_cases} shows images generated by our \methodNAME-T2I model, showcasing \methodNAME's strength in generating high-fidelity and photo-realistic images across various categories and styles. We also carry out the quantitative evaluation on the GenEval\cite{ghosh2024geneval} and DPG\cite{DPG-bench} benchmarks. As in Tab.\ref{tab:genEvalSotaTable}, \methodNAME achieves the best overall score of 0.79 on the GenEval bench with a prompt rewriter. It's worth noting that \methodNAME exceeds Infinity by 6\% on overall score. We attribute the significant improvement to the larger model size and the architectural innovations.  On the DPG bench, \methodNAME reaches an overall score of 86.55, surpassing Infinity by 3.09\%. These quantitative results demonstrate \methodNAME's strong capabilities of image generation following users' prompts.

\begin{table*}[t]
\centering
\caption{Evaluation on the GenEval~\cite{ghosh2024geneval} and DPG~\cite{DPG-bench} benchmark. $\dagger$ result is with prompt rewriting or self-CoT.}
\label{tab:genEvalSotaTable}
\resizebox{0.9\linewidth}{!}{ 
\begin{tabular}{lcccccccc}
        \toprule
    	\multirow{2}{*}{Methods} & \multirow{2}{*}{\# Params} & \multicolumn{4}{c}{GenEval$\uparrow$} & \multicolumn{3}{c}{DPG$\uparrow$} \\\cmidrule(l){3-6}\cmidrule(l){7-9}
        & & Two Obj. & Position & Color Attri. & \textbf{Overall} & Global & Relation & \textbf{Overall} \\
    	\midrule
            \multicolumn{9}{l}{Diffusion Models} \\
            \midrule
            SDXL~\citep{sdxl} & {2.6B} & 0.74 & 0.15 & 0.23 &  0.55 &    83.27   &   86.76   & 74.7 \\
            PixArt-Sigma~\cite{chen2024pixart_sigma} & 0.6B & 0.62 & 0.14 & 0.27 & 0.55 &   86.89   & 86.59 &   80.5 \\
            SD3 (d=38)~\citep{stable-diffusion3} & {8B}  &  \underline{0.89} & 0.34 & 0.47 & 0.71 & - & - &  - \\
            Goku~\citep{goku} & {2B}  &  - & - & - & 0.76$^{\dagger}$ & - & - &  83.6 \\
            Transfusion~\citep{transfusion} & {7.3B} & - & - & - & 0.63 & - &- & - \\
            SANA-1.0~\citep{xie2024sana} & {1.6B}  &  - & - & - & 0.66 & - & - &  84.8 \\
            FLUX-dev~\citep{FLUX} & {12B}  &  - & - & - & 0.67 & - & - &  84.0 \\
            FLUX-schnell~\citep{FLUX} & {12B}  &  - & - & - & 0.71 & - & - &  84.8 \\
            \midrule
            \multicolumn{9}{l}{AutoRegressive Models} \\
            \midrule
            LlamaGen~\citep{llamagen} & 0.8B & 0.34 & 0.07 & 0.04 & 0.32 &  &   &   65.2 \\
            Chameleon~\citep{chameleon-meta} & {7B} & - & - & - & 0.39 & - &- & - \\
            Show-o~\citep{show-o} & 1.3B & 0.80 &  0.31 &  0.50 & 0.68 & - &- & 67.5 \\
            Liquid~\citep{liquid} & 7B & 0.73 &  0.17 &  0.37 & 0.55 & - &- & - \\
            UniTok~\citep{unitok} & 7B & 0.71 &  0.26 &  0.45 & 0.59 & - &- & - \\
            Janus~\citep{janus} & 1.3B & 0.68 &  0.46 &  0.42 & 0.61 & - &- & - \\
            Emu3~\cite{wang2024emu3} & {8B} & 0.81$^{\dagger}$ & \underline{0.49$^{\dagger}$} & 0.45$^{\dagger}$ & 0.66$^{\dagger}$ & - & - & 81.6 \\
            Fluid~\cite{fluid} & 10.5B & 0.83 & 0.39  & 0.51 & 0.69 & - &- & - \\
            NextStep-1~\cite{nextstep1} & 14B & - & -  & - & 0.73$^{\dagger}$ & - &- & 85.28 \\
            Infinity~\cite{Infinity} & {2B} & 0.85$^{\dagger}$ & \underline{0.49$^{\dagger}$} & \underline{0.57$^{\dagger}$} & \underline{0.73$^{\dagger}$} & \textbf{93.11} & \underline{90.76} & \underline{83.46} \\
            \cellcolor{mycolor_green}{\textbf{\methodNAME-T2I}} & \cellcolor{mycolor_green}{8B} & \cellcolor{mycolor_green}{\textbf{0.90}$^{\dagger}$} & \cellcolor{mycolor_green}{\textbf{0.62}$^{\dagger}$} & \cellcolor{mycolor_green}{\textbf{0.67}$^{\dagger}$} & \cellcolor{mycolor_green}{\textbf{0.79}$^{\dagger}$} & \underline{\cellcolor{mycolor_green}{91.68}} & \cellcolor{mycolor_green}{\textbf{91.87}} & \cellcolor{mycolor_green}{\textbf{86.55}}\\
        \bottomrule
\end{tabular}
}
\end{table*}

\begin{table*}[t]
    \vspace{-0.4cm}
    \centering
    \caption{Evaluation on the VBench benchmark. $\dagger$ result is with prompt rewriting.}
    \label{tab:vBenchTable}
    \resizebox{0.95\linewidth}{!}{ 
    \begin{tabular}{lcccccccc}
   \toprule
        \multirow{2}{*}{Models} & \multirow{2}{*}{\# Params}  & Human  & \multirow{2}{*}{Scene} & Multiple  & Appear.  & Quality & Semantic & \multirow{2}{*}{\textbf{Overall}}  \\ 
         &  & Action &  & Objects & Style  & Score & Score &  \\
        \midrule
        \multicolumn{9}{l}{Diffusion Models} \\
        \midrule
AnimateDiff-V2 & 1.5B & 92.60 & 50.19 & 36.88 & 22.42 & 82.90 & 69.75 & 80.27 \\
VideoCrafter-2.0\citep{videocrafter} & 1.5B & 95.00 & \underline{55.29} & 40.66 & \textbf{25.13} & 82.20 & 73.42 & 80.44 \\
OpenSora V1.2\citep{opensora} & 1.1B & 85.80 & 42.47 & 58.41 & 23.89 & 80.71 & 73.30 & 79.23 \\
Show-1\citep{show-1} & 6B & 95.60 & 47.03 & 45.47 & 23.06 & 80.42 & 72.98 & 78.93 \\
Gen-3~\cite{Gen-3} & - & 96.40 & 54.57 & 53.64 & 24.31 & 84.11 & 75.17 & 82.32 \\
CogVideoX-5B\citep{cogvideox} & 5B & \textbf{99.40} & 53.20 & 62.11 & \underline{24.91} & 82.75 & 77.04 & 81.61 \\
HunyuanVideo\citep{Hunyuanvideo} & 13B & 94.40 & {53.88} & 68.55 & 19.80 & 85.09 & 75.82 & 83.24 \\
Goku\citep{goku} & 2B & 97.60 & \textbf{57.08} & \underline{79.48} & 23.08 & \underline{85.60} & \textbf{81.87} & \textbf{84.85} \\
Wan 2.1\citep{Wan} & 14B & \underline{98.80} & 53.67 & \textbf{81.44} & 21.13 & \textbf{85.64} & \underline{80.95} & \underline{84.70} \\
        \midrule
        \multicolumn{9}{l}{AutoRegressive Models} \\
        \midrule
Nova\citep{nova}$\dagger$ & 0.6B & 95.20 & 54.06 & 77.52 & 20.92 & 80.39 & 79.05 & 80.12 \\
Emu3\citep{wang2024emu3} & 8B & 77.71 & 37.11 & 44.64 & 20.92 & 84.09 & 68.43 & 80.96 \\
\cellcolor{mycolor_green}{\textbf{\methodNAME}$\dagger$} & \cellcolor{mycolor_green}{8B} & \cellcolor{mycolor_green}{96.43} & \cellcolor{mycolor_green}{52.08} & \cellcolor{mycolor_green}{\underline{78.66}} & \cellcolor{mycolor_green}{21.81} & \cellcolor{mycolor_green}{\underline{84.73}} & \cellcolor{mycolor_green}{\underline{79.78}} & \cellcolor{mycolor_green}{\underline{83.74}} \\

\bottomrule
\vspace{-0.7cm}
\end{tabular}
}
\end{table*}

\subsection{Text-to-Video Generation}
In the lower part of Fig.\ref{fig:show_cases}, we present the generated videos of \methodNAME regarding user prompts. The generated videos successfully capture the semantic information in user prompts while maintaining high aesthetics and visual quality. Especially for the second example in Fig.\ref{fig:show_cases}, the generated video accurately restores the delicate movements of the characters flipping through sketchbooks, talking while pointing to different parts of the drawings. In Tab.\ref{tab:vBenchTable}, we compare \methodNAME with leading diffusion and autoregressive approaches on VBench—a comprehensive video benchmark spanning 16 evaluation dimensions. Our model achieves an overall score of 83.74, outperforming all open-source autoregressive baselines by a substantial margin. Moreover, \methodNAME surpasses diffusion-based competitors such as OpenSora\cite{opensora}, CogVideoX\cite{cogvideox}, and HunyuanVideo\cite{Hunyuanvideo}. These results demonstrate that, through its novel spacetime autoregressive design, \methodNAME not only pushes the capabilities of discrete autoregressive video models but also attains performance on par with—and in some cases superior to—state-of-the-art diffusion methods.

\begin{figure}[t]
  \centering
   \includegraphics[width=1\linewidth]{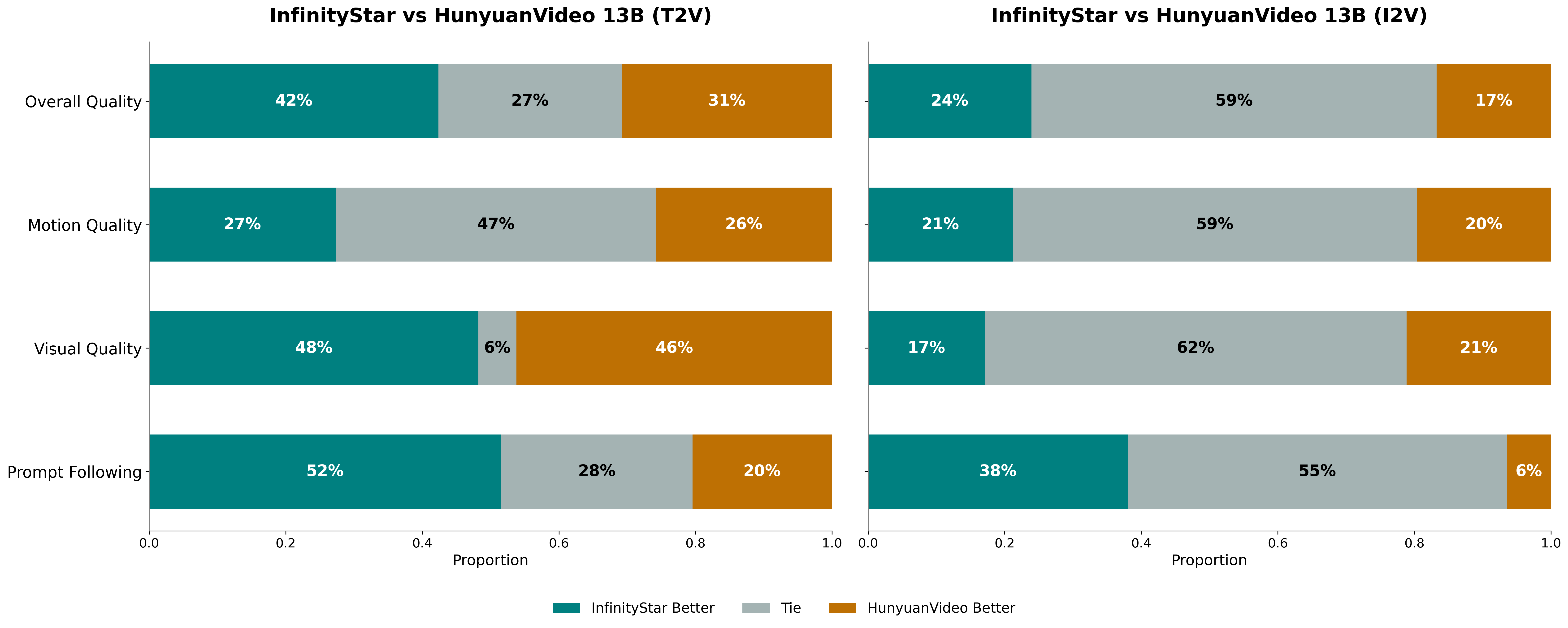}
   \vspace{-0.45cm}
   \caption{Human evaluation results comparing our model with HunyuanVideo 13B.}
   \label{fig:huamn_evaluation}
\end{figure}

\textbf{Human Preference Evaluation.} We conduct comprehensive human evaluation to benchmark our unified model, InfinityStar-8B, against a leading diffusion competitor, HunyuanVideo-13B. Specifically, we compared InfinityStar-8B to both the T2V and I2V variants of HunyuanVideo-13B. In a side-by-side comparison format, human raters were presented with videos generated by our model and those from HunyuanVideo-13B, and asked to judge which video was superior. Fig.\ref{fig:huamn_evaluation} lists the results of two human preference benchmarks. For the T2V task, our model consistently outperformed HunyuanVideo-13B across all evaluation metrics, even while maintaining a notable speed advantage. For the I2V task, InfinityStar-8B also demonstrated superior performance, particularly in prompt following and overall quality, compared to HunyuanVideo-13B. These results highlight the robust capability of InfinityStar 8B in generating high-quality videos that adhere closely to textual prompts.

\begin{figure}[t]
  \centering
   \includegraphics[width=1\linewidth]{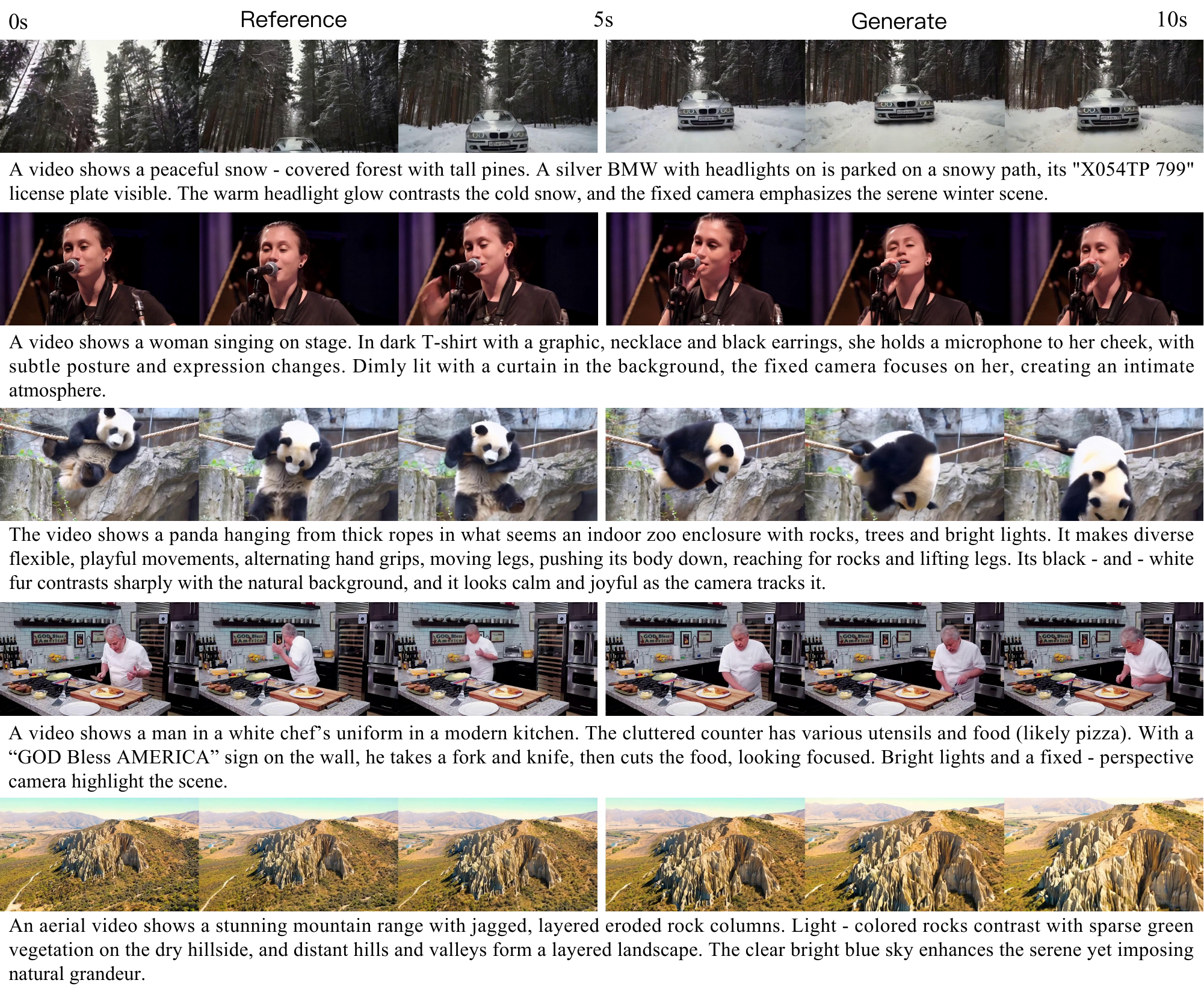}
   \vspace{-0.6cm}
   \caption{Zero-shot video extrapolation examples. \methodNAME can extrapolate videos using a reference video as historical without any fine-tuning.}
   \label{fig:video_extrapolation}
\end{figure}

\textbf{Zero-shot Generation.} Although trained exclusively on T2V data, \methodNAME can generate videos conditioned on an image or a video as historical without any fine-tuning. Fig.\ref{fig:video_extrapolation} shows video extrapolation results. The synthesized videos exhibit strong temporal coherence with the reference while faithfully capturing the semantic nuances of texts. Zero-shot I2V samples are presented in the appendix~\ref{appendix:C}.

\begin{table}[t]
\small
\centering
\caption{Reconstruction metrics on an internal high-motion video benchmark (480p 81 frames).}
\label{tab:high-motion}
\vspace{-0.2cm}
\begin{tabular}{lrrr}
  \toprule
  Pretrained Weights      & PSNR($\uparrow$) & SSIM($\uparrow$) & LPIPS($\downarrow$) \\
  \midrule
  Continuous Video VAE    & \textbf{33.37}   & \textbf{0.94}    & \textbf{0.065}      \\
  Image VAE               & 29.10            & 0.90             & 0.123               \\
  None                    & 30.04            & 0.90             & 0.124               \\
  \bottomrule
\end{tabular}
\end{table}

\vspace{-0.2cm}
\begin{figure}[t]
  \centering
   \includegraphics[width=1\linewidth]{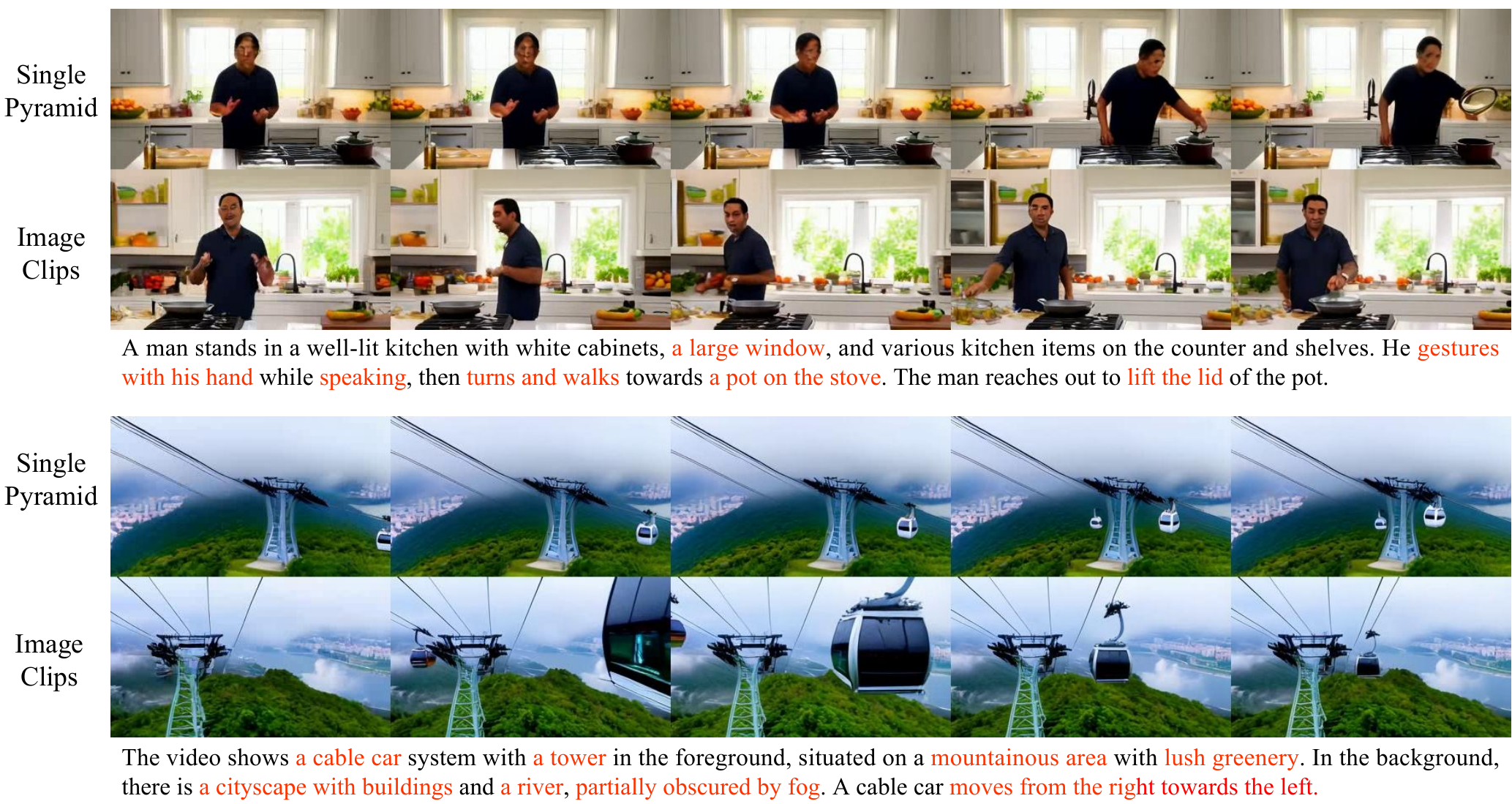}
   \vspace{-0.3cm}
   \caption{Comparison between Pseudo-Spacetime Pyramid and Spacetime Pyramid. Spacetime Pyramid could generate videos with richer details and higher motion.}
   \label{fig:ablation_pyramids}
   \vspace{-0.2cm}
\end{figure}

\begin{figure}[t]
  \centering
   \includegraphics[width=1\linewidth]{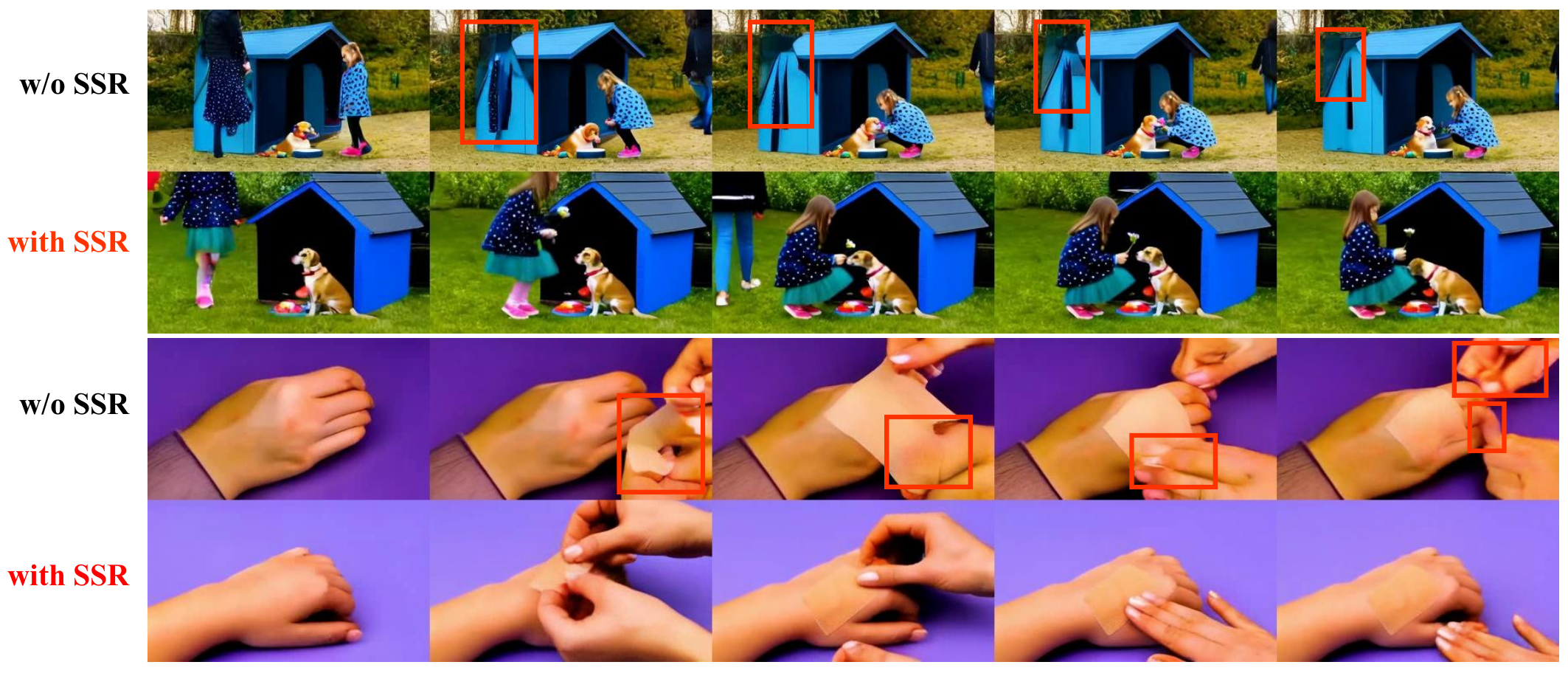}
   \vspace{-0.35cm}
   \caption{Semantic Scale Repetition (SSR) greatly improves structure stability and motion quality.}
   \label{fig:ablation_boost}
   \vspace{-0.1cm}
\end{figure}

\subsection{Ablation Study}

\textbf{Visual Tokenizer.} As shown in Fig.\ref{fig:recon-compare} and Tab.\ref{tab:high-motion}, loading weights of continuous video tokenizer significantly speeds up the convergence and achieves the best reconstruction results. As shown in Fig.\ref{fig:sqd_recon_vis}, stochastic quantizer depth largely improves the reconstruction quality of early scales. In terms of generation, using tokenizer with SQD leads to an improvement of 0.21 in VBench scores (81.28 \emph{v.s.} 81.07 as shown in Tab.\ref{tab:ablation_study_multirow}). Moreover, we observe that SQD contributes to faster convergence during the video generation training.

\textbf{Pseudo-Spacetime Pyramid \emph{v.s.} Spacetime Pyramid.} 

As illustrated in Fig.\ref{fig:ablation_pyramids}, videos generated by the pseudo-spacetime pyramid lack visual details and deliver simpler motion. In contrast, spacetime pyramid generates videos with richer details and higher motion. Besides, spacetime pyramid improves VBench's overall score from 80.30 to 81.28 as illustrated in Tab.\ref{tab:ablation_study_multirow}. These experiments support the hypothesis that spacetime pyramid could decouple appearance and temporal information. The image pyramid corresponds to the appearance information and clip pyramids focus on subsequent motions. This decoupling makes it easier to learn motions in videos. In addition to advances in performance, spacetime pyramid unifies T2I, T2V, I2V tasks into one framework.

\textbf{Semantic Scale Repetition.} In Fig.\ref{fig:sqd_recon_vis}, we can observe that the earlier scales correspond to semantic information, while the latter ones are responsible for high-frequency details. Here we compare the generation results of with and without semantic scale repetition. As shown in Fig.\ref{fig:ablation_boost}, semantic scale repetition is highly effective in improving the structure stability and motion quality. The quantitative results further confirm the significant gains. As shown in Tab.\ref{tab:ablation_study_multirow}, semantic scale repetition improves VBench's overall score from 75.72 to 81.28.

\textbf{Spacetime Sparse Attention.} In Tab.\ref{tab:ablation_study_multirow} and Tab.\ref{tab:ssa_efficiency}, we compare different attention mechanisms. 
Spacetime sparse attention shows superior performance to full attention in the Vbench total score (81.28 \emph{v.s.} 80.77), while showing a significant advantage in saving computation and GPU VRAM. SSA reaches 1.5$\times$
 speedup when generating 192p 161 frames. The efficiency advantage becomes larger as the resolution and duration grow. For 480p 161 frames, full attention fails due to OOM while SSA completes it within 44.7s using 63GB VRAM. We hypothesize that SSA produces better results than full attention because it reduces exposure bias. Full attention is more susceptible to accumulated errors. The reason we do not condition on smaller scales of the preceding clip is that it misses the former clips' visual details and brings visual inconsistency between clips. Although it reaches 1.1$\times$
, 1.5$\times$ speedup for 192p and 480p 161 frames, we observe a significant performance drop in Vbench from 81.28 to 80.75 as shown in Tab.\ref{tab:ablation_study_multirow}. Therefore, the proposed spacetime sparse attention strikes a better balance between computational efficiency and visual quality.

\subsection{Inferency Latency}
As shown in Tab.\ref{tab:inference_latency}, we report the end-to-end inference latency measured on a single GPU, including both the text encoder and VAE decoder. Wan-2.1\citep{Wan} and Nova\citep{nova} were evaluated using their default GitHub configurations. Even without employing stronger compression, \methodNAME achieves a 32$\times$ speedup over Wan-2.1. Furthermore, despite its 13$\times$ larger model size, \methodNAME delivers a 6$\times$ speedup compared to Nova. These results highlight our model’s significant advantage in efficiency over both diffusion and autoregressive approaches.

\begin{table}[h!]
\centering
\caption{Comprehensive ablation studies. Experiment with 1M 192p training data, batch size of 40, and 30K iterations. We evaluate the results on the Vbench benchmark.}
\label{tab:ablation_study_multirow}
 \resizebox{0.86\linewidth}{!}{
\begin{tabular}{lrrr}
  \toprule
  \multirow{2}{*}{Vbench} & total & quality & semantic \\ 
  & score & score & score \\ 
  \midrule
  \textbf{InfinityStar (Our Model)} & \textbf{81.28} & \textbf{81.56} & 80.16 \\ 
 \textit{Attend to former clip's largest scale} & & & \\
  \midrule
  \multicolumn{4}{l}{\textit{Ablation by removing/replacing core components}} \\
  w/o Semantic Scale Repetition(SSR) & 75.72 & 76.73 & 71.68 \\ 
  w/o Spacetime Pyramid (using Pseudo-Spacetime) & 80.30 & 80.81 & 78.28 \\ 
  w/o Stochastic Quantizer Depth(SQD) & 81.07 & 81.21 & \textbf{80.54} \\ 
  \midrule
  \multicolumn{4}{l}{\textit{Comparison of different Attention Mechanism variants}} \\
  Full Attention & 80.77 & 81.15 & 79.23 \\ 
  Attend to former clip's 3rd largest scale & 80.86 & 81.26 & 79.26 \\ 
  Attend to former clip's 6th largest scale & 80.75 & 80.98 & 79.80 \\ 
  \bottomrule
\end{tabular}
}
\end{table}

\begin{table}[h!]
\small
\centering
\caption{Computational efficiency comparison of attention mechanisms on a single GPU.}
\label{tab:ssa_efficiency}
\begin{tabular}{lrrr}
  \toprule
  & (192p 65 frames) & (192p 161 frames)  & (480p 161 frames) \\ 
  \midrule
  Full Attention &  8.6s / 40.8GB & 24.3s / 57GB & OOM \\ 
  Attend to former clip's largest scale &  7.7s / 38.5GB & 16.7s / 40GB & 44.7s / 63 GB \\ 
  Attend to former clip's 3rd largest scale  & 7.4s / 38.2GB & 15.8s / 39GB & 34.5s / 58 GB \\ 
  Attend to former clip's 6th largest scale & {7.3s / 37.9GB} & {15.2s / 38GB} & {30.5s / 55GB} \\ 
  \bottomrule
\end{tabular}
\vspace{-0.1cm}
\end{table}

\begin{table}[h!]
\centering
\caption{Computational efficiency comparison.}
\label{tab:inference_latency}
\vspace{-0.2cm}
\resizebox{1.0\linewidth}{!}{
\begin{tabular}{lccccccc}
\toprule
Method & Model & \# Parameters & Durations(s) & Frames & Resolution & Time(s) & Speedup\\
\midrule
Diffusion & Wan 2.1\citep{Wan} & 14B & 5 & 81 & 720p & 1864 & 1 \\
AR & Nova\citep{nova} & 0.6B & 5 & 81 & 480p & 354 & 5 \\
AR & \methodNAME & 8B & 5 & 81 & 720p & 58 & 32\\
\bottomrule
\end{tabular}
}
\vspace{-0.1cm}
\end{table}

\section{Extended Application: Long Interactive Video Generation}
The long interactive  video generation task focuses on the collaborative generation between the T2V model and users, accepting new user instructions and generating corresponding video content iteratively. While the original \methodNAME supports generating 10-second 480p videos, it only accepts a single prompt input and is limited to two clips. Extrapolating to longer video lengths than training involves performance degradation due to the discrepancy between training and testing. Simply increasing the number of training clips will lead to excessively long training sequences, which in turn causes an OOM issue. Below we introduce the innovations to extend \methodNAME to support long interactive video generation.

\begin{figure}[t]
  \centering
   \includegraphics[width=1.0\linewidth]{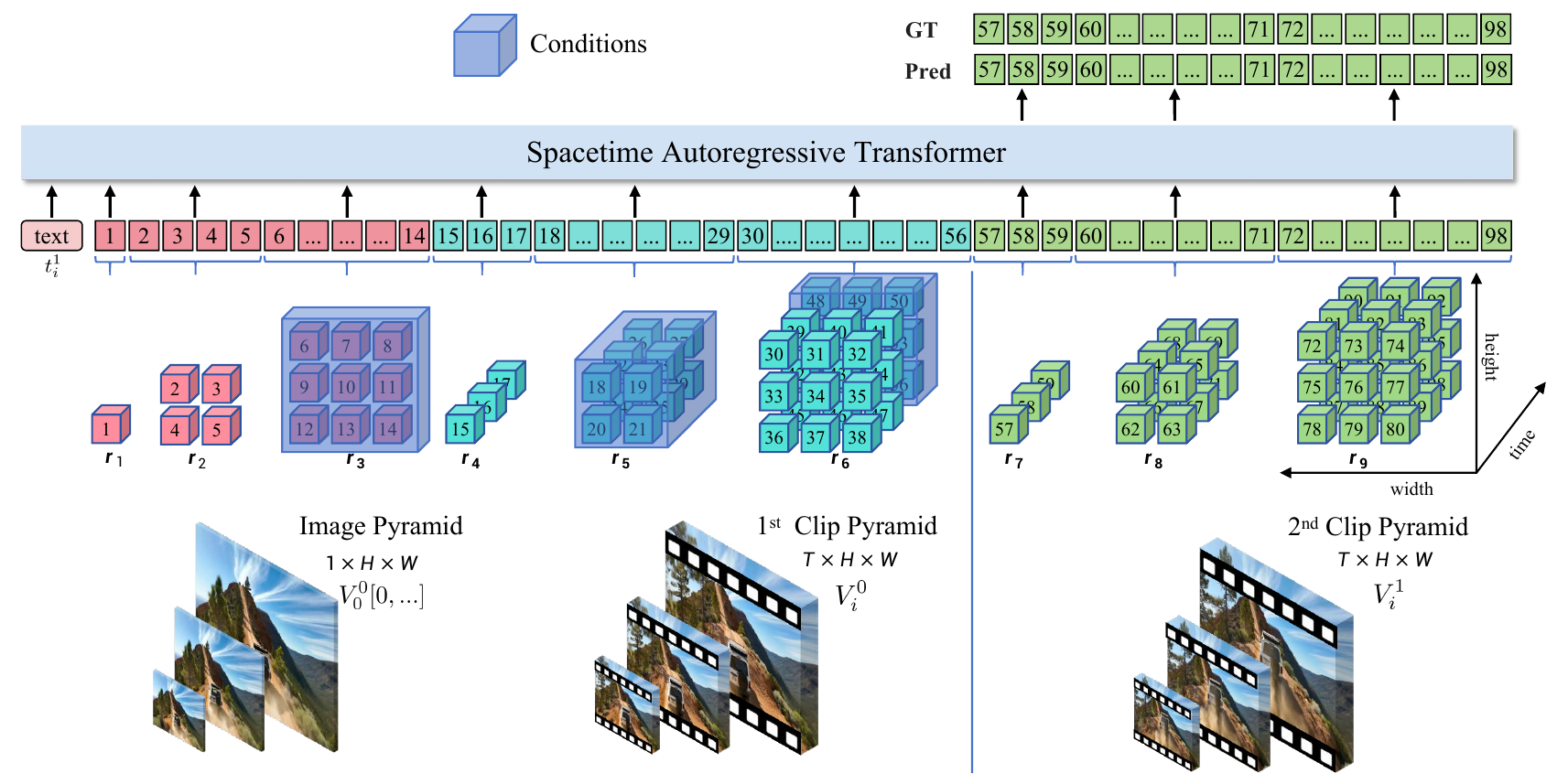}
   \vspace{-0.5cm}
   \caption{\textbf{Framework of \methodNAME-Interact.} We propose Semantic-Detail conditions (illustrated in light blue cubes) to control video synthesizing when interacting with users. It delivers superior visual and semantic consistency, as well as strong prompt-following capabilities.}
   \label{fig:framework_interactive}
\end{figure}

\subsection{Model Design}
We solve the problem of long interactive video generation using a sliding window method. Mathematically, for a long interactive video $V \in T^{long} \times H \times W$, we decompose it into a series of video chunks of 10 seconds, \emph{i.e.}, $\{V_0, V_1, ..., V_n\}$, with stride of 5 seconds. Each chunk $V_i$ is further divided into two clips $V^0_i$ and $V^1_i$. Each video clip is 5 seconds long and paired with a transition caption, \emph{i.e.}, $t^1_{i-1}$ or $t^0_{i}$,  with the assistance of an LLM. Note that $(t^0_{i}, V^0_i)$ is the same with $(t^1_{i-1}, V^1_{i-1})$. During each round interaction with the user, \methodNAME generates $V^1_i$ conditioned on  $(V^0_0[0,...], V^0_i, t^1_{i})$, where $V^0_i$ is $V^1_{i-1}$ that we generated in the preceding interaction round. $V^0_0[0,...]$ is the first frame of the earlist video clip. This division method allows training on only two clips, while enabling to synthesize infinitely long videos during the inference stage. We find that conditioning on $V^0_0[0,...]$ could mitigate drift when generating multi-round videos. 

Beyond spacetime sparse attention, we introduce the novel Semantic-Detail conditions to control video synthesizing when interacting with users as illustrated in Fig.\ref{fig:framework_interactive}. Specifically, we extract features $F_{i-1} \in T\times H\times W$ from the preceding clip $V^1_{i-1}$ using the visual tokenizer. The features $F_{i-1}$ are referred to detail features since they are full-scale and contain rich visual details. It is difficult to extract semantic information from $F_{i-1}$ because it is not adequately compressed. Besides, there are too many tokens in $F_{i-1}$, which significantly slows down the interactive inference speed. Borrow ideas from FramePack \cite{framepack}, we downsample $F_{i-1}$ to $F_{i-1}^{sem}\in T\times h\times w$ spatially to reduce excessive condition tokens. The semantic conditions $F_{i-1}^{sem}$ are employed to enable semantic consistency between clips. Apart from $F_{i-1}^{sem}$, we slice the last $K$ frames from $F_{i-1}$ instead of the whole as the detail conditions $F_{i-1}^{det}\in K\times H\times W$. In this way, we ensure consistency in both semantics and details while significantly compressing the number of condition tokens.

\begin{figure}[t]
  \centering
   \includegraphics[width=0.8\linewidth]{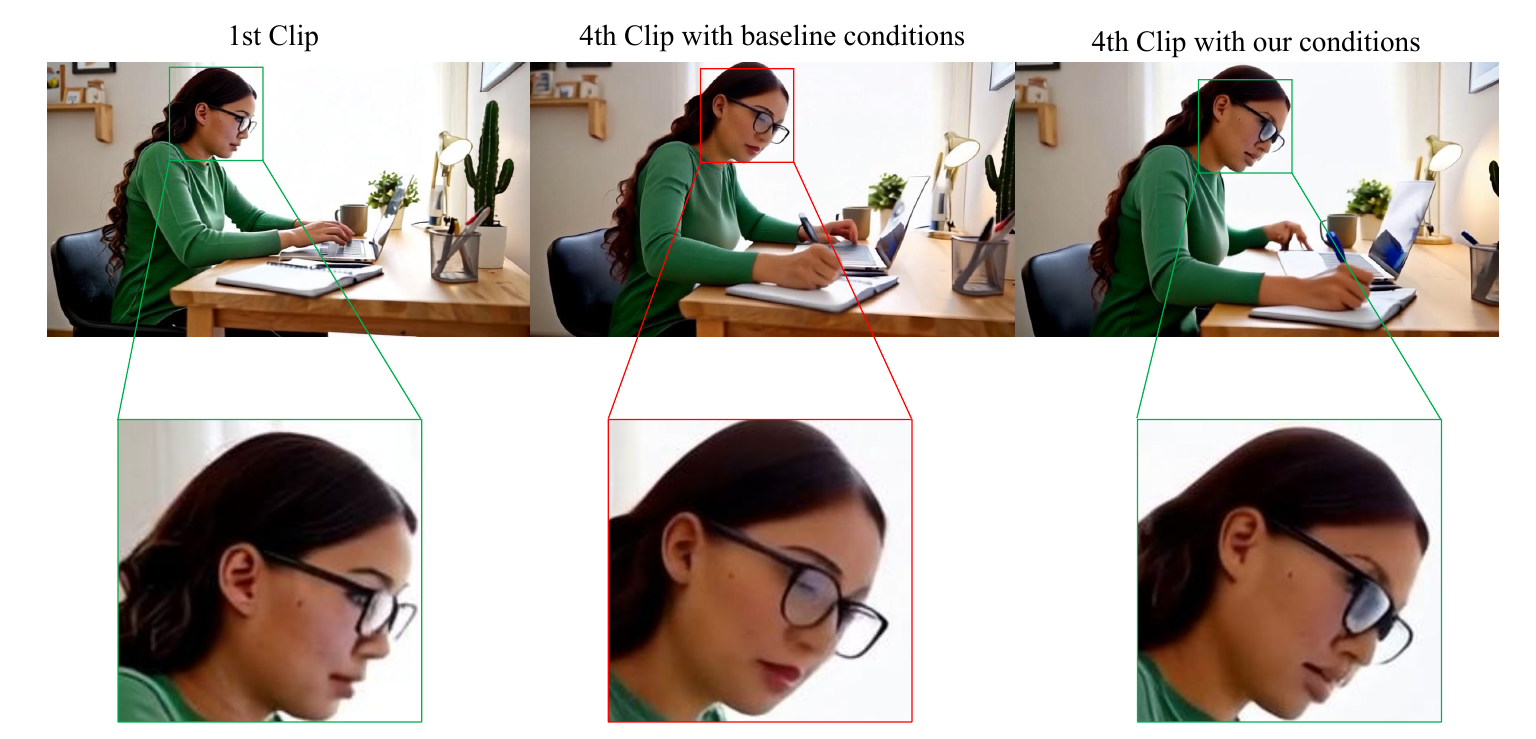}
   \vspace{-0.3cm}
   \caption{Conditioning solely on the last few frames of preceding clip (baseline conditions) is inadequate for preserving semantic consistency. Our proposed conditions deliver better capability in maintaining semantic consistency.}
   \label{fig:condition_compare}
\end{figure}

\begin{figure}[t]
  \centering
   \includegraphics[width=1.0\linewidth]{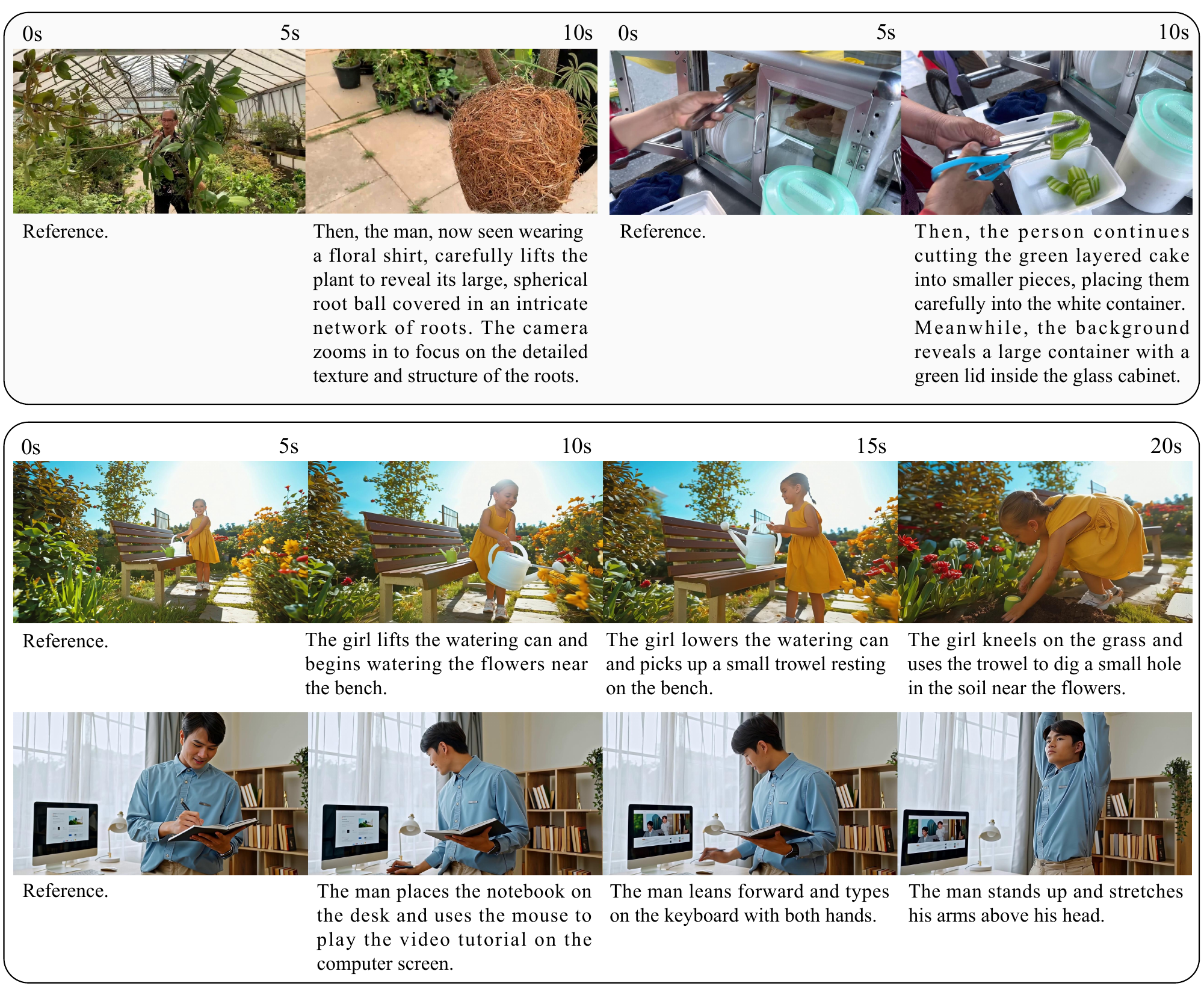}
   \vspace{-0.6cm}
   \caption{Examples of curated interactive training data. The upper part is obtained by selecting data from pre-training datasets and rewriting captions using an LLM. The lower part is synthetic interaction data, generated by first using an LLM to create prompts and then calling a video continuation model.}
   \label{fig:interactive_data}
\end{figure}

\begin{figure}[h]
  \centering
   \includegraphics[width=1.0\linewidth]{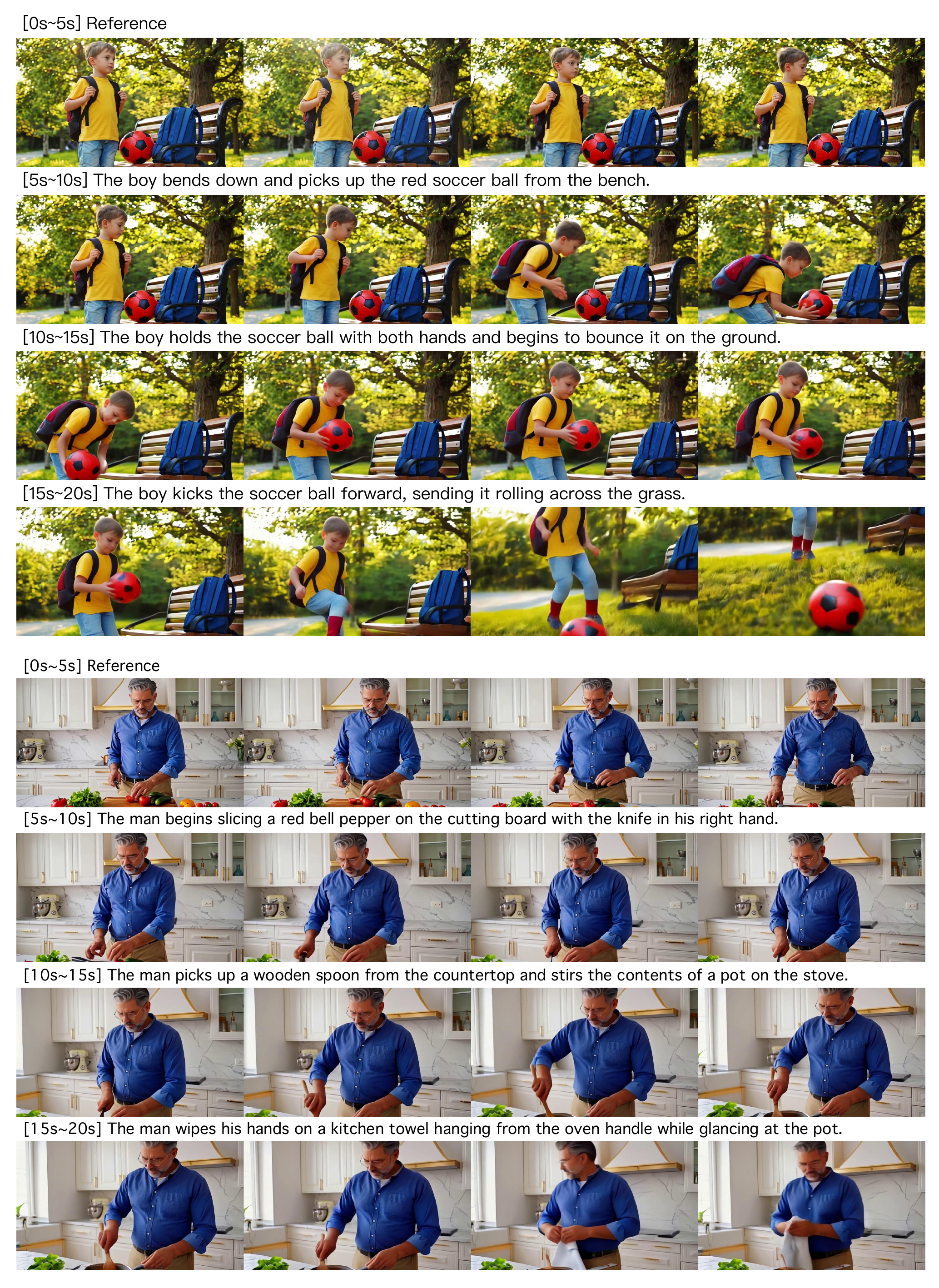}
   \vspace{-0.7cm}
   \caption{Interactive Generation Results. Given the first 5-second video as a reference, \methodNAME-Interact generates 480p videos through multi-round collaboration with users. Whether focusing on outdoor character movements (as in the first example) or indoor character hand movements (as in the second example), \methodNAME-Interact can generate interactive videos that follow users' prompts.}
   \label{fig:interactive_demo}
\end{figure}

\subsection{Dataset}
We curate the interactive generation data from the pretraining dataset and other sources. In particular, we select videos longer than 7 seconds from the pretraining data, resulting in a total of 7M videos. Subsequently, we decompose long videos into chunks, split the chunks into clips, and generate captions at the clip level using the Tarsier2 \cite{tarsier2} model. It is worth noting that here we adopt an LLM to remove the content that had already appeared in $V^0_i$'s caption from  $V^1_i$'s caption, and ensure that $t^1_i$ only describes changes compared to $t^0_i$. The instructions used to query an LLM are presented in the Appendix \ref{sub:instructions}. In this way, we align with the instructions users provide during interactive generation.

Apart from filtering pretraining data, we also incorporate some synthetic long interaction data. Specifically, we first collect multi-round interactive prompts. These prompts are used as seeds to query an LLM to generate more samples. We pick good ones from the generated samples to enlarge the seed set and query an LLM again to enhance diversity. Finally, we collect 16K interactive prompts, where each prompt is consists of four round interactions. Then we use the prompts to query a video continuation model to generate interaction videos. We provide the instrucitons to generate multi-round interactive prompts in the Appendix \ref{sub:instructions}.
We present some examples of the curated interaction data in Fig.\ref{fig:interactive_data}.

\subsection{Evaluation}
The training of the interactive generation model is divided into two stages. In the first stage, we load the weights of \methodNAME and conduct continued pre-training on the filtered pre-training data. The learning rate during this stage is set to 2e-4. In the second stage, we fine-tune the model on the synthetic interaction data. We decay the learning rate to 2e-5.
We slice the last 2 frames (set $K=2$) from the preceding clip as detail features. The semantic features are obtained by downsampling the detail features with a stride of $\sqrt{32}$. Compared to spacetime sparse attention, the proposed semantic-detail conditions compress the condition token length from 33.6K to 5.8K for 480P video generation.

Empirical observations reveal that relying solely on the last few frames of the preceding clip (abbreviated as baseline conditions) is inadequate to preserve semantic consistency in the long interactive generation task. Our proposed semantic-detail conditions deliver higher quality and better consistency in semantics while showing high efficiency. As shown in Fig.\ref{fig:condition_compare}, the face ID of the woman has changed after three rounds of interactive generation, whereas the proposed conditions have successfully maintained its consistency. Fig.\ref{fig:interactive_demo} presents two examples of \methodNAME-Interact. Whether outdoor character movements as in the first example or indoor character hand movements as in the second example,  \methodNAME-Interact generates consistent videos during interactions with the user.

\section{Conclusion}
\label{sec:conclusion}

\vspace{-0.3cm}
We introduce \methodNAME, a unified spacetime autoregressive framework capable of synthesizing high-resolution images and dynamic, high-motion videos. By seamlessly integrating spatial and temporal prediction within a purely discrete architecture, \methodNAME supports diverse generation tasks while maintaining both state-of-the-art quality and exceptional efficiency. Our extensive evaluation demonstrates that \methodNAME outperforms prior autoregressive video models and rivals leading diffusion-based approaches, producing a 720p video of 5s in one-tenth the inference time. Besides, we extend \methodNAME to support long interactive video generaiton. As the first discrete autoregressive model to deliver industrial-grade 720p video synthesis, we anticipate that \methodNAME will catalyze future research on rapid, long video generation.

\section{Limitation}
\label{sec:limitation}
While \methodNAME sets a new record in discrete video generation and demonstrates strong prompt following ability as well as impressive motion capabilities, several limitations remain. Specifically, there is a trade-off between image quality and motion fidelity in high-motion scenes, where sometimes fine-grained visual details can be compromised. Additionally, due to limited computational resources, we have not scaled our model training or parameter size to match those of leading diffusion models, which constrains the upper bound of the performance. Furthermore, our inference pipeline has not yet been fully optimized, indicating room for future improvement. In terms of the limitations in long interactive video generation, \methodNAME suffers from cumulative errors. With the increase in the number of interactions, there will be a noticeable degradation in the quality of the generated videos. This constitutes a problem that we are required to address.

\section{Acknowledgments}
The authors appreciate the valuable support provided by colleagues from ByteDance, including Yuqi Zhang, Yifu Zhang, Hao Yang, Yifei Hu, Chuang Lin, Xiaofeng Mei, Ruibiao Lu, and Jiawei Duan. Their contributions to data processing and related technical aspects are essential for the advancement of this research.

\clearpage
{
\small
\bibliographystyle{cite}
\bibliography{cite}
}


\appendix

\newpage




\section{Spacetime Autogressive Modeling}
\label{appendix:A}
\begin{figure}[h]
  \centering
   \includegraphics[width=0.95\linewidth]{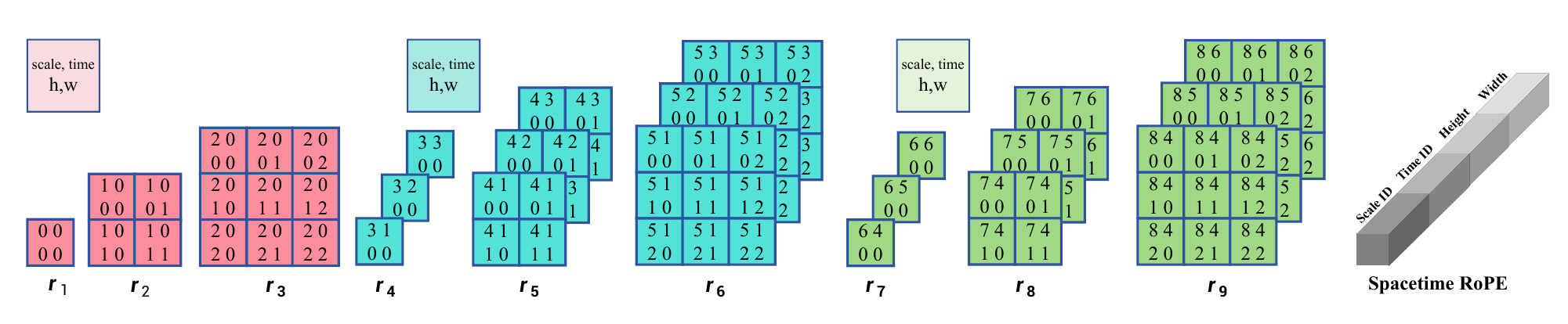}
   \vspace{-0.5cm}
   \caption{An illustration of Spacetime RoPE. We decompose rotary embeddings into four components, \emph{i.e.}, scale, time, height, and width components. Spacetime RoPE enhances the modeling of complex positional information while supporting extrapolation.}
   \label{fig:spacetime_rope}
   \vspace{-0.2cm}
\end{figure}

\noindent\textbf{Spacetime RoPE.} We introduce spacetime rotary position embeddings (Spacetime RoPE) tailored for \methodNAME. This is achieved by decomposing original rotary embeddings\cite{ROPE} into four components: scale, time, height, and width. As shown in Fig.\ref{fig:spacetime_rope}, the scale ID serves as a counter of scales up to now. The temporal ID remains zero for tokens in the image pyramid. For tokens in video pyramids, it increments as the frame grows. Distinct IDs are assigned to height and width components based on the token's position in the image or video. Spacetime RoPE enhances the modeling of complex positional information for tokens in image and video pyramids.

\noindent\textbf{Spacetime Autoregressive Transformer with Bitwise Self-Correction.}
To alleviate the train-test discrepancies of teacher-forcing training, we adopt bitwise self-correction mechanism proposed by Infinity\cite{Infinity}. Specifically, during training, some of the input tokens are randomly flipped to simulate the prediction error during the inference phase. Besides, the target labels are also recomputed to match the perturbed inputs. Moreover, when predicting the token distribution, the traditional index-wise classifier is replaced by a bitwise classifier. The bitwise classifier predicts $d$ bits instead of $2^d$ indices, significantly reducing the memory costs and difficulties in optimization. Algorithm 1 shows the detailed procedure of Spacetime Pyramid Encoding with Bitwise Self-Correction.

\begin{algorithm}[H]
\caption{Spacetime Pyramid Encoding with BSC} \label{alg:self_correction}
\begin{algorithmic}[0]
\Require raw feature $\bm{F}$,  scale schedule number $K$, clip number $N$\\
image pyramid scale schedule: $(1,h_1,w_1),\ldots,(1,h_{K},w_{K})$,  \\
clip  pyramid scale schedule: $(T,h_1,w_1),\ldots, (T,h_{K},w_{K})$  
\State $\bm{R}_{queue} \gets []$ \Comment{multi-scale bit labels}
\State $\widetilde{\bm{F}}_{queue} \gets []$ \Comment{inputs for transformer}
\For {$c=1,2,\ldots,N$}   \Comment{inter-clips iteration}
\State $t_{start} = 1 + (c-1) * T$ 
\State $\bm{F}_{c} \gets \text{raw features from time  } t_{start} \text{  to  } t_{start} + t_c$ 
\For {$k=1,2,\ldots,K$}   \Comment{intra-clip multi-scale iteration}
    \State $\bm{R}_k = \operatorname{quant}(\operatorname{down} (\bm{F}_{c} - \bm{F}^{flip}_{c,k-1}, (t_k, h_k, w_k))$
    \State $\operatorname{Queue\_Push}(\bm{R}_{queue}, \bm{R}_k)$
    \State $\bm{R}^{flip}_k = \operatorname{Random\_Flip}(\bm{R}_k, p)$
    \State $\bm{F}^{flip}_{c,k} = \sum_{i=1}^{k} \operatorname{up}(\bm{R}^{flip}_i, (h, w))$
    \State $\widetilde{\bm{F}}_{c,k} = \operatorname{down}(\bm{F}^{flip}_{c,k}, (t_{k+1}, h_{k+1}, w_{k+1}))$
    \State $\operatorname{Queue\_Push}(\widetilde{\bm{F}}_{queue}, \widetilde{\bm{F}}_{c,k})$
\EndFor
\EndFor
\Ensure $\bm{R}_{queue}$, $\widetilde{\bm{F}}_{queue}$
\end{algorithmic}
\end{algorithm}

\begin{figure}[t]
  \centering
   \includegraphics[width=1.0\linewidth]{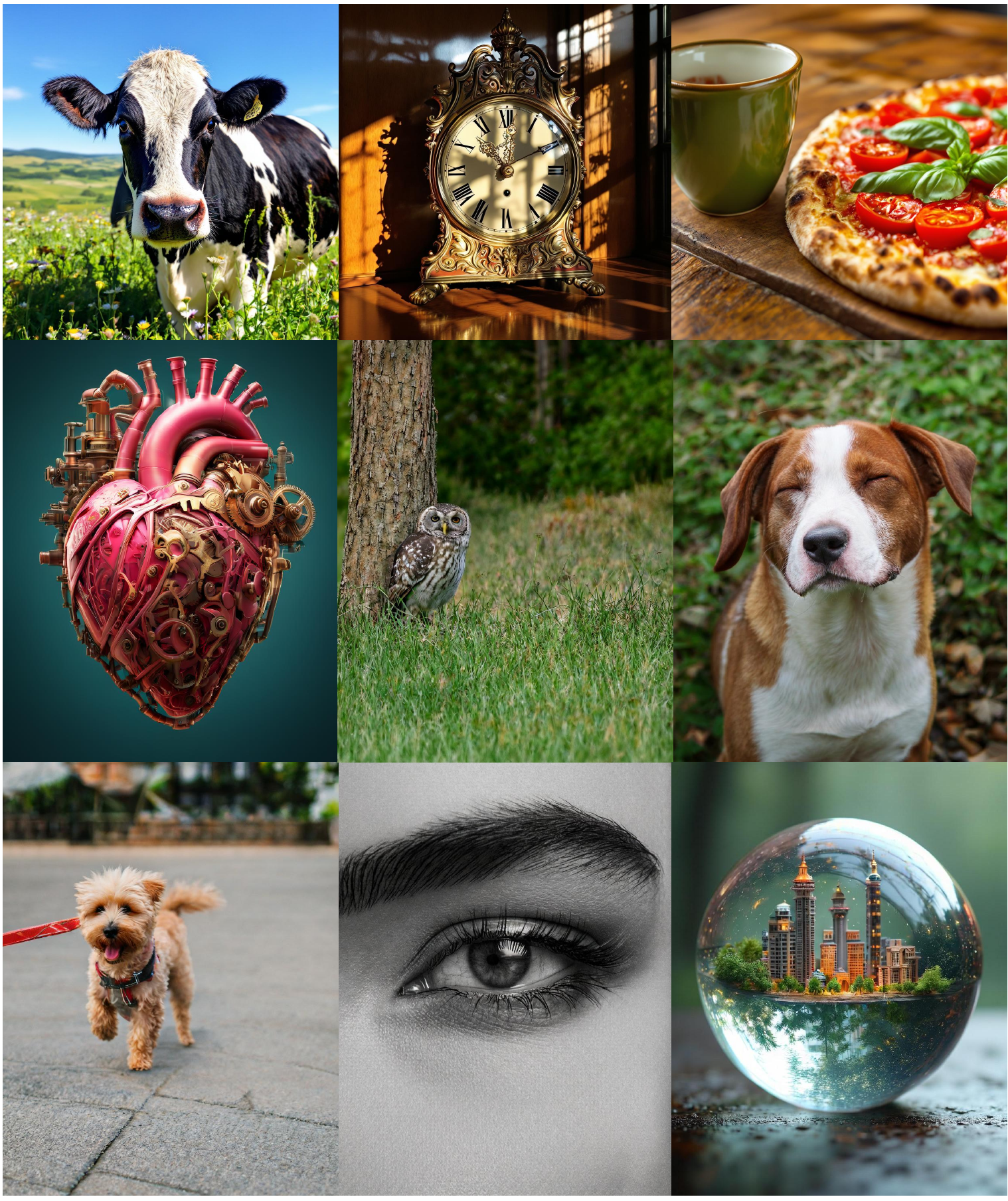}
   \vspace{-0.6cm}
   \caption{Text to image examples.}
   \vspace{-0.5cm}
   \label{fig:supp_show_cases}
\end{figure}

\section{Infrastructure and Data}
\label{appendix:B}
\textbf{Infrastructure Optimization.} Compared to diffusion models, visual autoregressive methods possess around 2.5$\times$ longer training sequences. This feature poses crucial pressure on hardware and algorithms when scaling models and increasing resolutions. In this work, we adopt advanced parallelism methods for scalable and efficient training. 

Firstly, we utilize FlexAttention to implement various attention mechanisms. With our proposed Spacetime Sparse Attention, we achieve more than a $2\times$ acceleration in training speed. Secondly, we adopt fully sharded data parallelism (FSDP) \cite{fsdp} to partition parameters, gradients, and optimizer states across GPU ranks. Thirdly, we adopt a fine-grained activation checkpointing strategy to reduce the overhead of vRAM and data transfer, making the parallelization more efficient. Last but not least, sequence parallelism further partitions long sequences into multiple chunks and then exploits ring self-attention for each chunk, making it feasible to train 720p videos with 200K sequence length.

\textbf{Visual Captioning.} Detailed visual captioning is crucial for enabling the model to accurately generate images and videos. For images, we use InternVL2.0\cite{chen2023internvl} to produce dense descriptions for each sample. For video clips, we obtain overall video descriptions using Tarsier2\cite{tarsier2}. Notably, Tarsier2 inherently captures camera motion types (e.g., zoom, pan right), eliminating the need for a separate prediction model. This simplifies the pipeline and improves efficiency.

\textbf{Data Pipeline.} Obtaining a high-quality image and video dataset requires a complex processing pipeline. Specifically for video, we follow video processing pipelines\cite{goku} to preprocess videos into high-quality training clips through OCR filtering, video clip extraction, visual aesthetic filtering, and motion filtering, etc.

\section{More Qualitative Results}
\label{appendix:C}

\subsection{Text-to-Image Generation.}
Fig.\ref{fig:supp_show_cases} shows more generated images from our \methodNAME-T2I model. Our model is capable of generating high-resolution images filled with vivid and intricate details.

\subsection{Zero-shot Generation}

\begin{figure}[t]
  \centering
  \vspace{-0.3cm}
   \includegraphics[width=1.\linewidth]{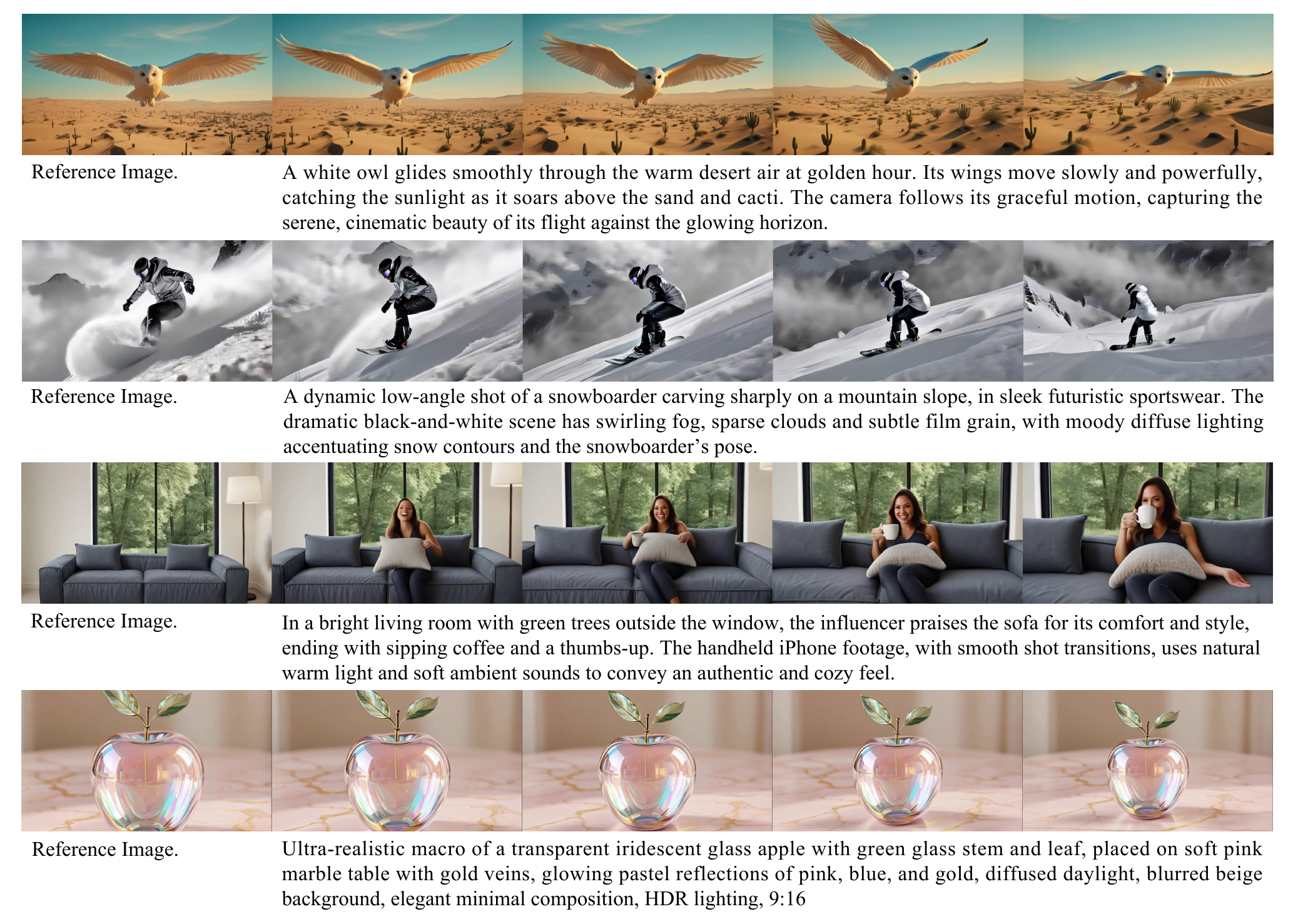}
   \vspace{-0.8cm}
   \caption{Zero-shot image to video examples. \methodNAME can generate videos following an input image without fine-tuning. The synthesized videos exhibit strong temporal and semantic coherence.}
   \vspace{-0.2cm}
   \label{fig:image2video}
\end{figure}

\noindent\textbf{Image to Video.}  Although trained exclusively on text-to-video data, \methodNAME can generate videos conditioned on an input image without any fine-tuning. Fig.\ref{fig:image2video} illustrates qualitative results on the image-to-video task. The synthesized videos exhibit strong temporal coherence with the reference image—a critical requirement for this task—while faithfully capturing the semantic nuances of the accompanying text with high visual fidelity. 

\begin{figure}[t]
  \centering
  \vspace{-0.2cm}
   \includegraphics[width=1.\linewidth]{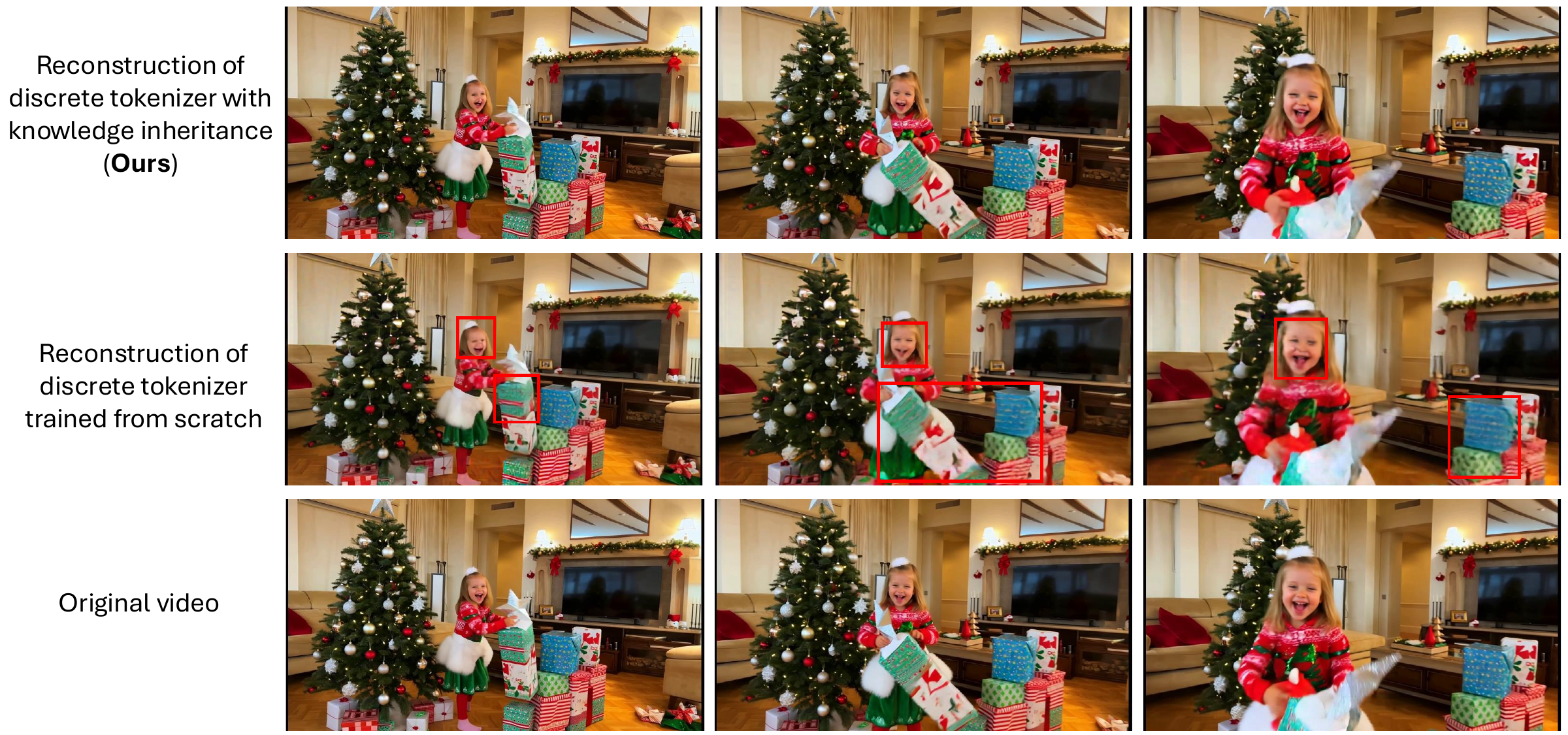}
   \vspace{-0.7cm}
   \caption{Comparison between the reconstruction quality of different video tokenizers. The tokenizer incorporating knowledge inheritance (top row) demonstrates a substantial improvement compared to one trained from scratch (middle row).}
   \label{fig:vae_spp}
\end{figure}

\subsection{Video Reconstruction.} Figure~\ref{fig:vae_spp} illustrates a comparison between the reconstructed videos generated by different tokenizers and the original video. The discrete tokenizer trained from scratch (middle row) exhibits inferior reconstruction quality. In contrast, the tokenizer incorporating knowledge inheritance (top row) demonstrates a substantial improvement in visual fidelity, particularly in the preservation of intricate details such as human faces and complex textures.

\subsection{Instructions.} 
\label{sub:instructions}

Below is the instruction for removing duplicate captions from adjacent clips.

\begin{small}
\begin{verbatim}
You are a helpful assistant. 
Paragraph 1: <<<clip 1's tarsier2 caption>>>
Paragraph 2: <<<clip 2's tarsier2 caption>>>
These two paragraphs describe a 10-second video: the first paragraph covers the first 
5 seconds, while the second focuses on the last 5 seconds. 
However, the second paragraph was written without considering the content already
included in the first one, resulting in significant repetition. 
Now, I need you to revise the second paragraph: 
• Remove the repetitive content that has already been mentioned in the first paragraph 
and retain only the new information. 
• You can think of the revised second paragraph as a description of what changes occurred 
in the last 5 seconds compared to the first 5 seconds. 
• If necessary, add sequential transition words such as "then" or "next" to better 
describe the changes.
• If no obvious differences are identified, you may first extract the core content from the 
previous paragraph and then add transition words like "continue" or "keep" to indicate continuity. 
• Please provide an analysis first, followed by the revised result. 
• Please place the revised results between "<<<" and ">>>"
\end{verbatim}
\end{small}

Below is the instruction for generating multi-round interactive
prompts.

\begin{small}
\begin{verbatim}
You are an expert in writing prompts. The written prompts are used to query a text-to
-video model to generate videos interactively. Each video is 20 seconds long and consists 
of four 5-second shots. Each shot shows the next moment of the same scene compared to 
the previous shot. For each new shot, you add a new action to the main subject from the 
previous shot. Describe the facts directly and do not use rhetoric. To prevent hallucinations, 
the objects in the subsequent three shots must have appeared in the first shot.
Below are some examples you have written before:
Example 1
<story>
<shot1>A young boy wearing a green hoodie and jeans is in a backyard with a wooden fence 
and green grass. A red ball, a blue bicycle, and a yellow toy truck are on the grass nearby. 
The boy is standing next to the red ball, looking at it with his hands on his hips.</shot1>
<shot2>The boy picks up the red ball with both hands.</shot2>
<shot3>The boy throws the red ball forward across the grass.</shot3>
<shot4>The boy runs toward the blue bicycle parked near the fence.</shot4>
</story>
Example 2
<story>
<shot1>A woman wearing a red sweater and glasses stands in a kitchen with white cabinets 
and a marble countertop. On the countertop are a cutting board with chopped vegetables, 
a stainless steel knife, a glass bowl, and a bottle of olive oil. The woman holds the 
knife in her right hand and is about to chop a tomato on the cutting board.</shot1>
<shot2>The woman finishes chopping the tomato and places the knife down on the cutting board.</shot2>
<shot3>The woman picks up the glass bowl and transfers the chopped vegetables into it.</shot3>
<shot4>The woman picks up the bottle of olive oil and pours some into the glass bowl.</shot4>
</story>
Example 3
<story>
<shot1>A man wearing a blue button-up shirt and black trousers stands in a small home 
office. The room contains a wooden bookshelf filled with books, a black swivel chair, 
and a desk with a desktop computer, a white coffee mug, and a closed notebook. The man 
holds a smartphone in his right hand, looking at the screen with a neutral expression.</shot1>
<shot2>The man puts the smartphone down on the desk next to the coffee mug.</shot2>
<shot3>The man sits down on the black swivel chair and opens the notebook on the desk.</shot3>
<shot4>The man picks up a pen from the desk and begins writing in the notebook.</shot4>
</story>
Please write three new examples and output them in the same format as the example. 
Don't be too similar to the written examples.
\end{verbatim}
\end{small}

\end{document}